\documentclass[10pt,twocolumn,letterpaper]{article}

\usepackage{cvpr}              

\usepackage{graphicx}
\usepackage{amsmath}
\usepackage{amssymb}
\usepackage{booktabs}
\usepackage{multicol}
\usepackage{multirow}
\usepackage{color}
\usepackage{arydshln}
\usepackage[table]{xcolor}
\usepackage{makecell}
\usepackage{algorithm}
\usepackage{algorithmicx}
\usepackage{algpseudocode}

\usepackage[pagebackref,breaklinks,colorlinks]{hyperref}

\usepackage[capitalize]{cleveref}
\crefname{section}{Sec.}{Secs.}
\Crefname{section}{Section}{Sections}
\Crefname{table}{Table}{Tables}
\crefname{table}{Tab.}{Tabs.}
\newcommand{\tcb}{\textcolor{black}}

\def\cvprPaperID{2650} 
\def\confName{CVPR}
\def\confYear{2023}
\definecolor{mygray}{gray}{.1}
\newcommand{\tcr}{\textcolor{black}}

\begin{document}

\title{Learning Distortion Invariant Representation for Image Restoration \\ from A Causality Perspective}

\author{Xin Li\textsuperscript{\rm 1}\footnotemark[1], Bingchen Li\textsuperscript{\rm 1}\footnotemark[1], Xin Jin\textsuperscript{\rm 2}, Cuiling Lan\textsuperscript{\rm 3}\footnotemark[2], Zhibo Chen\textsuperscript{\rm 1}\footnotemark[2]\\
\textsuperscript{\rm 1}University of Science and Technology of China \quad
\textsuperscript{\rm 2}Eastern Institute for Advanced Study \\
\textsuperscript{\rm 3}Microsoft Research Asia \\
\tt\small \{lixin666, lbc31415926\}@mail.ustc.edu.cn, jinxin@eias.ac.cn, \\ \tt\small culan@microsoft.com,  chenzhibo@ustc.edu.cn}
\maketitle
\renewcommand{\thefootnote}{\fnsymbol{footnote}}
\footnotetext[1]{Equal contribution}
\footnotetext[2]{Corresponding Author}
\begin{abstract}
In recent years, we have witnessed the great advancement of Deep neural networks (DNNs) in image restoration. 
However, a critical limitation is that they cannot generalize well to real-world degradations with different degrees or types. 
In this paper, we are the first to propose a novel training strategy for image restoration from the causality perspective, to improve the generalization ability of DNNs for unknown degradations.
Our method, termed \textbf{D}istortion \textbf{I}nvariant representation \textbf{L}earning (DIL), treats each distortion type and degree as one specific confounder, and learns the distortion-invariant representation by eliminating the harmful confounding \tcb{effect} of each degradation. We derive our \textbf{DIL} with the back-door criterion in causality by modeling the interventions of different distortions from the optimization perspective. 
Particularly, we introduce the counterfactual distortion augmentation to simulate the virtual distortion types and degrees as the confounders. Then, \tcr{we instantiate the intervention of each distortion with a virtually model updating based on corresponding distorted images, and eliminate them from the meta-learning perspective.}
Extensive experiments demonstrate the effectiveness of our \textbf{DIL} on the generalization capability on unseen distortion types and degrees. Our code will  be available at \url{https://github.com/lixinustc/causal-IR-DIL}.
\end{abstract}

\section{Introduction}
\label{sec:intro}
\begin{figure}
    \centering
    \includegraphics[width=0.90\linewidth]{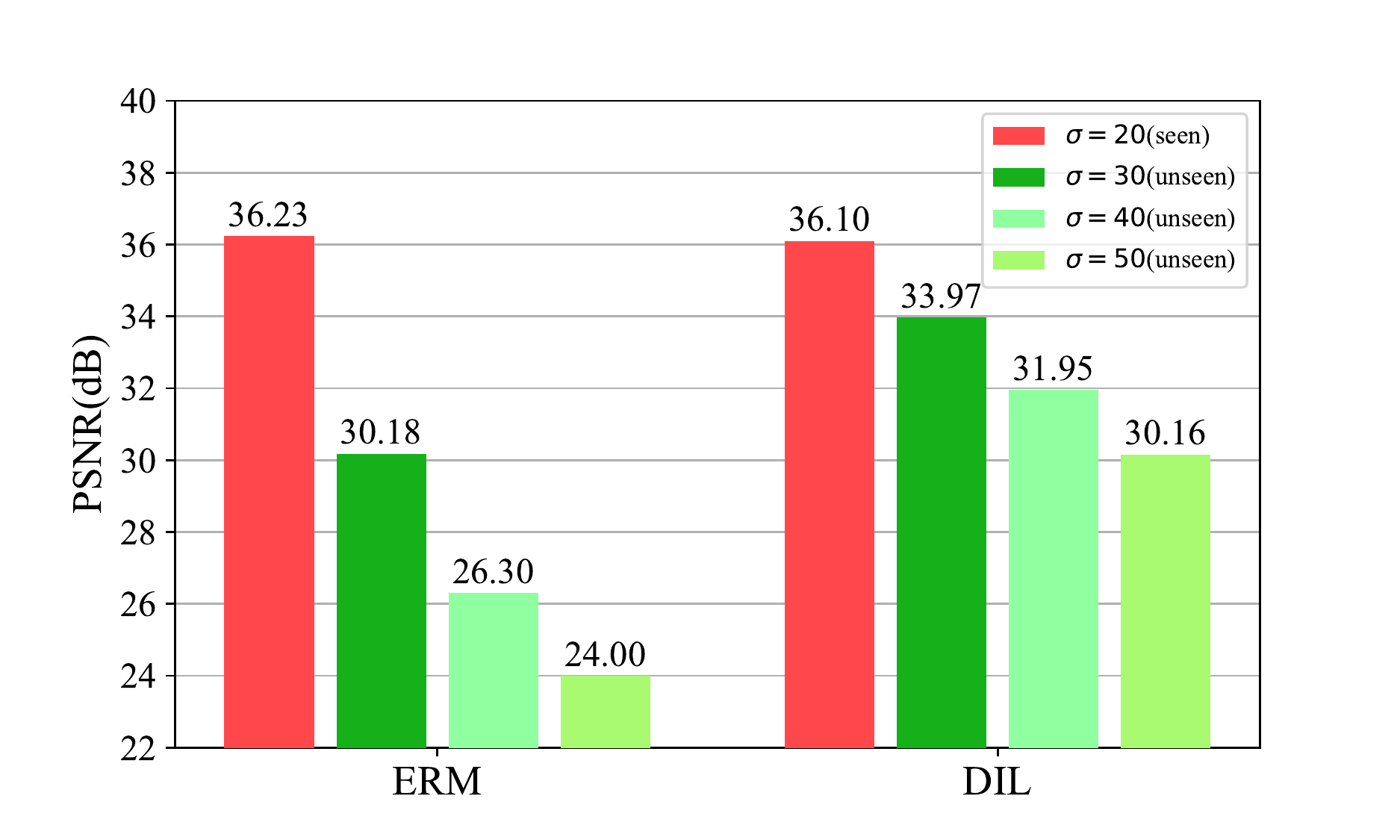}
    \caption{A comparison between ERM and our \textbf{DIL} with RRDB as backbone. The results are tested on the Set5 with gaussian noise.}
    \label{fig:intro}
    \vspace{-6mm}
\end{figure}

Image restoration (IR) tasks~\cite{chen2021preIPT,liang2021swinir,tu2022maxim,liif}, including image super-resolution~\cite{ESRGAN,dong2015imageSRCNN,li2022hst,MZSR,park2020fast,wei2020aim}, deblurring~\cite{nah2021cleanDblur,zhang2020residualRDN}, denoising~\cite{mohan2019robustdenoise,anwar2019realRIDNet,lee2020self}, compression artifacts removal~\cite{li2020multiMSGDN,wang2021jpeg}, etc, have achieved amazing/uplifting performances, powered by deep learning.
A series of backbones are elaborately and carefully designed to boost the restoration performances for specific degradation. Convolution neural networks (CNNs)~\cite{he2016deepResNet} and transformers~\cite{liu2021swin,dosovitskiy2020image} are two commonly-used designed choices for the backbones of image restoration. However, these works inevitably suffer from severe performance drops when they encounter unseen degradations as shown in Fig.~\ref{fig:intro}, where the restoration degree in training corresponds to the noise of \tcr{standard deviation 20} and the degrees in testing are different. The commonly-used training paradigm in image restoration, \textit{i.e.,} empirical risk minimization (ERM), does not work well for out-of-distribution degradations.
Particularly, the restoration networks trained with ERM merely mine the correlation between distorted image $I_d$ and its ideal reconstructed image $I_o$ by minimizing the distance between $I_o$ and the corresponding clean image $I_c$. However, a spurious correlation~\cite{pearl2009causal} is also captured which introduces the bad confounding effects of specific degradation $d$.
It means the conditional probability $P(I_o|I_d)$ is also conditioned on the distortion types or degrees $d$ (\textit{i.e.,} $d \not \! \perp \!\!\! \perp I_o|I_d$).

A robust/generalizable restoration network should be distortion-invariant (\textit{i.e.,} $d \! \perp \!\!\! \perp I_o|I_d$). For instance, given two distorted images with the same content $I_c$ but different degradations $d_1$ and $d_2$, the robust restoration network is expected to recover the same reconstructed image $I_o$ from these two distorted images (\textit{i.e.,} $P(I_o|I_d, d=d_1)=P(I_o|I_d, d=d_2)$), respectively. Previous works for the robustness of the restoration network can be roughly divided into two categories, distortion adaptation-based schemes, and domain adaptation/translation-based schemes. Distortion adaptation-based schemes~\cite{wei2021unsupervisedDASR} aim to estimate the distortion types or representations, and then, handle the different distortions by adaptively modulating the restoration network. Domain adaptation/translation-based 
schemes~\cite{sun2021learningZIZO,lugmayr2019unsupervisedURealSR,du2020learningLIRUIR} regard different distortions as different domains, and introduce the domain adaptation/translation strategies to the image restoration field. Notwithstanding, the above works ignore the exploration of the intrinsic reasons for the poor generalization capability of the restoration network. In this paper, we take the first step to the causality-inspired image restoration, where novel distortion invariant representation learning from the causality perspective is proposed, to improve the generalization capability of the restoration network.

\begin{figure}[htp]
    \centering
    \includegraphics[width=0.8\linewidth]{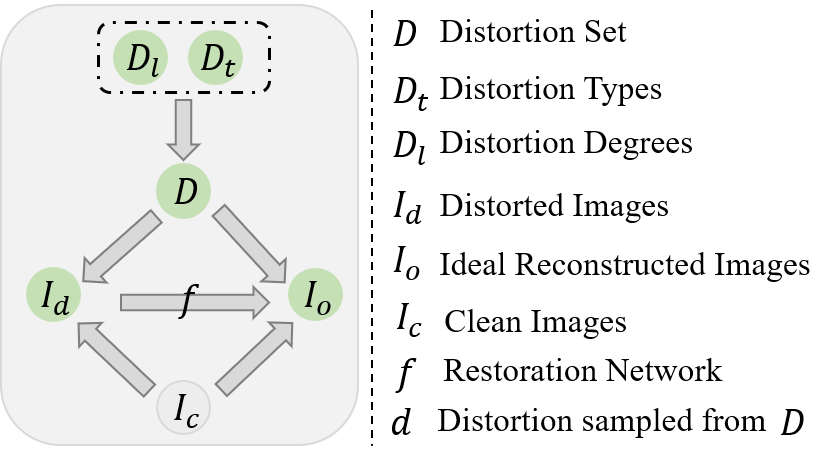}
    \caption{Causal structure graph for image restoration.}
    \label{fig:SCM_IR}
    \vspace{-5mm}
\end{figure}

As depicted in~\cite{pearl2009causal,glymour2016causality}, \textit{correlation is not equivalent to causation}. Learning distortion invariant representation for image restoration requires obtaining the causal effects between the distorted and ideal reconstructed images instead of only their correlation. From the causality perspective, we build a causal graph for the image restoration process. As shown in Fig.~\ref{fig:SCM_IR}, the distortions $D$ including types $D_t$ or degrees $D_l$ are the confounders in IR, \tcr{which introduces the harmful bias and causes the restoration process $I_d\xrightarrow[]{}I_o$ condition on $D$, }
since a spurious relation path is established via $I_d\xleftarrow[]{}D\xrightarrow[]{}I_o$. The causal connection we want between distorted and ideal reconstructed image is $I_d\xrightarrow[]{}I_o$, of which the causal conditional probability can be represented as $P(I_o|do(I_d))$. Here, a ``$do$" operation~\cite{pearl2009causal,glymour2016causality} is exploited to cut off the connection from the distortion $D$ to $I_d$, thereby removing the bad confounding effects of $D$ to the path $I_d\xrightarrow[]{}I_o$, and learning the distortion-invariant feature representation (\textit{i.e., $D \! \perp \!\!\! \perp I_o|I_d$}).  

There are two typical adjustment criteria for causal effects estimation~\cite{glymour2016causality}, the back-door criterion, and the front-door criterion, respectively. In particular, the back-door criterion aims to remove the bad confounding effects by traversing over known confounders, while the front-door criterion is to solve the challenge that confounders cannot be identified. To improve the generalization capability of the restoration network, we propose the \textbf{D}istortion-\textbf{I}nvariant representation \textbf{L}earning (DIL) for image restoration by implementing the back-door criterion from the optimization perspective. 
\tcr{There are two challenges for achieving this.  The first challenge is how to construct the confounder sets (\textit{i.e.,} distortion sets). From the causality perspective~\cite{pearl2009causal,glymour2016causality}, it is better to keep other factors in the distorted image invariant except for distortion types. However, in the real world, collecting the distorted/clean image pairs, especially with different real distortions but the same content is impractical.}
Inspired by counterfactual~\cite{pearl2009causal} in causality and the distortion simulation~\cite{RealESRGAN,BSRGAN}, 
we propose the counterfactual distortion augmentation, which 
 selects amounts of high-quality images from the commonly-used  dataset~\cite{DIV2K,Flickr2K}, and simulate the different distortion degrees or types on these images as the confounders. 
 
 
 
Another challenge of implementing DIL stems from finding a stable and proper instantiation scheme for back-door criterion. Previous works~\cite{lv2022causality,mahajan2021domain,yue2021transportingUDA,yang2021causal,wang2020visualVCR} have incorporated causal inference in high-level tasks by instantiating the back-door criterion~\cite{glymour2016causality} with attention intervention~\cite{yang2021causal}, feature interventions~\cite{yue2020interventionalIFL}, etc, which are arduous to be exploited in the low-level task of image restoration. 
In this work, we theoretically derive our distortion-invariant representation learning \textbf{DIL} by instantiating the back-door criterion from the optimization perspective. Particularly, we model the intervention of simulated distortions for the restoration process by virtually updating the restoration network with the samples from the corresponding distortions. Then, we eliminate the confounding effects of distortions by introducing the optimization strategy from Meta Learning to our proposed DIL. In this way, we can instantiate the causal learning in image restoration and \tcb{enable} the \textbf{DIL} based on the back-door criterion. 

The contributions of this paper are listed as follows:
\vspace{-3mm}
\begin{itemize}
    \item We revisit the image restoration task from a causality view and pinpoint that the reason for the poor generalization of the restoration network, is that the restoration network is not independent to the distortions in the training dataset. 
    \vspace{-2mm}
    \item Based on the back-door criterion in causality, we propose a novel training paradigm, \textbf{D}istortion \textbf{I}nvariant representation \textbf{L}earning (\textbf{DIL}) for image restoration, where the intervention is instantiated by a virtually model updating under the counterfactual distortion augmentation and is eliminated with the optimization based on meta-learning. 
    \vspace{-3mm}
    \item Extensive experiments on different image restoration tasks have demonstrated the effectiveness of our \textbf{DIL} for improving the generalization ability on unseen distortion types and degrees. 
\end{itemize}
\vspace{-4mm}
\section{Related Works}
\label{sec:related_work}
\vspace{-2mm}
\subsection{Image Restoration}
\vspace{-2mm}
Image Restoration (IR)~\cite{liang2021swinir,li2020learningFDRNet,chen2021preIPT,liu2020liraLIRA,tu2022maxim,jin2021dualIR,zamir2022restormer} aims to recover high-quality images from the corresponding distorted images, which plays a prominent role in improving the human visual experience. With the advancement of deep learning, a series of works have achieved remarkable progress in lots of IR tasks, including image denoising~\cite{anwar2019realRIDNet,zhang2022idrIDR,mohan2019robustdenoise}, deblurring~\cite{nah2021cleanDblur,tran2021explore,zhang2020residualRDN}, super-resolution (SR)~\cite{dong2015imageSRCNN,zhang2018imageRCAN,dai2019secondSAN,wei2020aimAIMRealSR,yang2022aimAIMCIR,li2021learningORNet}, etc.  Particularly, most of them are devoted to elaborately designing the frameworks for different IR tasks based on their degradation process, which can be roughly divided into two categories, CNN-based framework~\cite{dong2015imageSRCNN,zhang2018imageRCAN,dai2019secondSAN}, and Transformer-based framework~\cite{liang2021swinir,chen2021preIPT,li2022hst,zamir2022restormer}. Despite that, the above works only explore how to improve the ability of inductive bias toward specific degradation, which lacks enough generalization capability. 
To improve the model's robustness, some works seek to incorporate the domain translation~\cite{sun2021learningZIZO,lugmayr2019unsupervisedURealSR,du2020learningLIRUIR} or distortion-adaptive learning~\cite{wei2021unsupervisedDASR} into image restoration. In contrast, we introduce causal learning~\cite{glymour2016causality} to image restoration. We answer the reason for the bad robustness of the restoration network and propose \textbf{\textit{distortion-invariant representation learning from a causality perspective}}. 
\vspace{-3mm}
\subsection{Causal Inference}
\vspace{-2mm}
Causal Inference is proposed to eliminate the harmful bias of confounders and discover the causal relationship between multiple variables~\cite{glymour2016causality}. A $do$ operation is implemented with adjustment criteria, \textit{e.g.,} front-door or back-door, to estimate the causal effects~\cite{pearl2009causal}. In recent years, deep learning has boosted the vast development of a series of intelligent tasks, \textit{e.g.,} image classification~\cite{dosovitskiy2020image,deng2009imagenet,liu2021swin}, segmentation~\cite{he2017mask,strudel2021segmenter}, detection~\cite{carion2020end,li2022exploring}, low-level processing~\cite{chen2021preIPT,tu2022maxim}. However, prominent works focus on fitting the \textit{correlation} between inputs and their \tcb{outputs} while ignoring the \textit{causation}. Due to the existence of confounders, the networks are easy to  capture the spurious correlation between inputs and their \tcb{outputs}. For instance, if \textit{most lions lie in the grass} in the training data, the model inevitably mistakes the grass for a lion. To get rid of the harmful bias of confounders, some studies seek to incorporate causal inference into deep learning. \cite{yue2020interventionalIFL,wang2020visualVCR} model the interventions of confounders from the feature perspective
\cite{yang2021causal} integrate the front-door criterion to vision-language task from the attention perspective. To improve the generalization capability, \cite{yue2021transportingUDA,mahajan2021domain,li2021confounderCICF,lv2022causality} introduce the causal learning to domain adaptation/generalization. However, the above causality-inspired methods merely \tcb{focus on} the high-level tasks. In this paper, \tcb{for the first time}, we investigate the causality-based image restoration, which aims to improve the generalization capability of restoration networks on different distortion types and degrees. 

\section{Methods}
\vspace{-2mm}
\subsection{A Causal View \tcb{for} Image Restoration}
\vspace{-2mm}
Image restoration aims to restore the distorted images, of which the degradation process can be represented as a function $I_d = g(I_c, D)$. Here, $I_c, I_d, D$ denotes the clean, distorted images, and distortions, respectively. A restoration network $f$ is trained with the loss function to \tcb{minimize the difference} between its ideal reconstructed images $I_o$ and the original clean image $I_c$. We model this whole process with a causal structure graph as shown in Fig.~\ref{fig:SCM_IR}. Here, $D\xrightarrow[]{}I_d \xleftarrow[]{}I_c$ denotes the degradation process of $I_d = g(I_c, D)$.  $I_c \xrightarrow[]{}I_o$ denotes $I_o$ is learned with the supervision of $I_c$ by maximizing the probability of $P(I_c|I_o)$. 
\tcr{In addition, $D\xrightarrow[]{}I_o$ refers to the knowledge learned from $D$ to $I_o$.} $I_d\xrightarrow[]{}I_o$ means the restoration process with restoration network $f$. 

From the causality perspective, the causal representation of image restoration requires that the restoration network $f$ obtains the causal relationship between $I_d\xrightarrow[]{}I_o$ (\textit{i.e.}, $P(I_o|do(I_d))$). However, there are two extra paths $I_d\xleftarrow[]{}D\xrightarrow[]{}I_o$ and $I_d\xleftarrow[]{}I_c\xrightarrow[]{}I_o$ introducing the spurious correlation to $I_d$ and $I_o$, where $I_c$ and $D$ are confounders in causality. Importantly, the $I_c$ are commonly diverse in the datasets and bring more vivid textures to reconstructed image $I_o$, which is a favorable confounder. We do not take into account of the confounder $I_c$
in our paper.

We aim to improve the robustness of the restoration network to unseen or unknown distortions, which are inhibited by the bad confounding effects from confounder\tcb{s} $D$. 
But, \textit{how do the confounders $D$ limit the generalization capability of the restoration network?} As shown in Fig.~\ref{fig:SCM_IR}, the existing of $I_d\xleftarrow[]{}D\xrightarrow[]{}I_o$ causes the conditional probability $P(I_o|I_d)$ learned by restoration network $f$ is also condition on distortions $D$, \textit{i.e.,} the fitting conditional probability of $f$ is in fact as $P(I_o|I_d, D)$. Consequently, the restoration network $f$ is not robust to different distortions due to \tcb{that it is not independent of different distortions D. }

A robust restoration network $f$ should be independent of different distortions (\textit{i.e.,} $D\! \perp \!\!\! \perp I_o|I_d$). To achieve this, we adopt the back-door criterion in causal inference to realize distortion-invariant learning (DIL). We formulate the back-door criterion in image restoration as Equ.~\ref{equ:back-door}. 
\begin{equation}
    \centering
    P(I_o|do(I_d)) = \sum_{d_i\in D} P(I_o|I_d, d_i)P(d_i), P(d_i)=\frac{1}{n},
    \label{equ:back-door}
\end{equation}
where the causal conditional probability $P(I_o|do(I_d))$ is the optimization direction for restoration network $f$ towards distortion invariant learning. To simplify the optimization, we set the probability of each distortion $d_i$ as $1/n$, where $n$ is the number of distortion types and degrees that existed in confounders. From Equ.~\ref{equ:back-door}, two crucial challenges for achieving it arise.  1) \textit{How to construct the virtual confounders (\textit{i.e.,} different distortion types or degrees)? since collecting different real distorted images with the same contents are nontrivial in the real world.} 2) \textit{How to instantiate the intervention of different distortions to the reconstruction process (\textit{i.e.,} the $P(I_o|I_d, d_i)$) in image restoration}. We \tcb{achieve this through counterfactual distortion augmentation and distortion-invariant representation learning as described} in the following sections.

\subsection{Counterfactual Distortion Augmentation}
\tcr{To learn the distortion-invariant representation for the restoration network, it is vital to construct the distortion set $D$ (\textit{i.e.}, confounders). For instance, if we expect the restoration network to have the generalization capability for different distortion degrees, we require to construct the distortion set $D$ with the distortions at different levels. Similarly, we can increase  the generalization capability of the restoration network for unknown distortion types by constructing the $D$ with different but related distortion types. Furthermore, to avoid the effects of different image contents, it is better for each clean image to have corresponding distorted images with various distortion types or degrees in $D$. However, it is non-trivial to collect datasets that satisfy the above principles in the real world, which is labor-intensive and arduous.}

\tcr{In this paper, we construct the distortion set $D$ with synthesized distortions, which we can call them virtual confounders in causality. Concretely, we collected a series of high-quality images $I_c$, and generated the distorted images by modifying the degradation process as $I_d=g(I_c, d_i), d_i\in D$. We can also prove the rightness of the above distortion augmentation from the counterfactuals in causlity~\cite{glymour2016causality}, where we  answer the counterfactual question that \textit{``if $D$ is $d_i$, what the $I_d$ would be with $I_c$ invariant?". The proof can be found in the \textbf{Supplementary}.}}

\subsection{Distortion-invariant Representation Learning}
After constructing the virtual confounders/distortions set $D=\{d_i|1 \le i\le n\}$. We are able to achieve the distortion-invariant representation learning by implementing the back-door criterion as Eq.~\ref{equ:back-door} for image restoration. Let us first introduce the relationship between the probability $P(I_o|I_d)$ and the commonly-used training paradigm ERM (empirical risk minimization). In image restoration, an ideal reconstruction $I_o$ is expected to learn by maximizing the condition probability $P(I_o|I_d)$ with loss function as $\mathcal{L}(f_\theta(I_d), I_c)$, where $f_\theta$ is the restoration network with the parameters $\theta$ \tcb{and} $L$ denote\tcb{s} the loss function, such as the commonly-used $\mathcal{L}_1$ or $\mathcal{L}_2$ loss. The ERM is used to optimize the network $f_\theta$ \tcb{(with parameters denoted by $\theta$)} by minimizing the loss function overall training dataset $\mathcal{D}=\{I_d, I_c|d\in D\}$ as:
\begin{equation}
    \theta^* = \arg \min_{\theta} \mathbb{E}_{(I_d,I_c)\sim \mathcal{D}} [\mathcal{L} (f_{\theta}(I_d), I_c)],
\end{equation}
\tcr{where $\theta^*$ enables the restoration network $f$ to maximize the $P(I_o|I_d)\approx P(I_c|I_d)$. }
\tcr{However, the above training process also leads the $P(I_o|I_d)$ to be not independent to the distortions $d \in D$ in the training dataset $\mathcal{D}$, which eliminate the generalization ability of $f$ on the out-of-distribution distortions (\textit{i.e.,}, when $d \not \in D$). }
To achieve the distortion-invariant representation learning, we \tcb{aim} to maximize the causal conditional probability $P(I_o|do(I_d))$ \tcb{as} instead of $P(I_o|I_d)$. The \tcb{key} challenge stems from how to model the conditional probability $P(I_o|I_d, d_i)$ in Eq.~\ref{equ:back-door} (\textit{i.e.}, how to model the intervention from the distortion $d_i\in D$ for the restoration process $P(I_o|I_d)$).

In this paper, we propose to model the intervention  from $d_i\in D$ to the restoration process (\textit{i.e.,} $P(I_o|I_d, d_i)$) through the \tcb{optimization of the} network parameters $\theta$. 
\tcb{From} the above analysis, we know that the restoration network $f_\theta$ trained with ERM on the paired training \tcb{data} $(I_{d_i}, I_c)$ \tcb{is} condition on the distortion $d_i$. This inspires us to instantiate the intervention of different distortion types or degrees $d_i\in D$ through updating the model parameter $\theta$ to ${\phi_{d_i}}$ based on ERM with the training distorted\tcb{-and-}clean image pairs $(I_{d_i}, I_c)$ 
as:
\begin{equation}
    \centering
    \phi_{d_i} = \theta - \alpha \nabla_\theta \mathcal{L}(f_\theta(I_{d_i}), I_c),
    \label{equ:pre-update}
\end{equation}
where $\phi_{d_i}$ denotes the parameters of the restoration network \tcb{after one-step update}, \tcb{which} is condition\tcb{ed} on the confounder $d_i$. 
Consequently, the maximum of the conditional probability $P(I_o|I_d, d_i)$ can be obtained by minimizing the loss $\mathcal{L}(f_{\phi_{d_i}}(I_d), I_c)$.
The optimization direction toward maximizing the causal condition probability $P(I_o|do(I_d))$ in Eq.~\ref{equ:back-door} can be derived as:
\begin{equation}
    \theta^* = \arg \min_{\theta} \mathbb{E}_{(I_d, I_c)\sim \mathcal{D}}[ \sum_{d_i\in D} \mathcal{L}(f_{\phi_{d_i}}(I_d), I_c)],
    \label{equ:final-update}
\end{equation}
where $D$ denotes the confounder set which contains $n$ distortion degrees or types. Based on the above optimization \tcb{objective}, we \tcb{learn} distortion-invariant representation learning from a causality perspective.
\begin{table*}[ht]
\centering
\caption{Quantitative comparison for image denoising on several benchmark datasets. Results are tested on three different unseen distortion degrees in terms of PSNR/SSIM on RGB channel. Best performances are \textbf{bolded}.}
\resizebox{0.95\textwidth}{!}{\begin{tabular}{c|c|c|c|c|c|c} 
\hline
\multirow{2}{*}{Datasets} & \multirow{2}{*}{Levels} & \multicolumn{5}{c}{Methods}                                \\ 
\cline{3-7}
                          &                        & ERM          & DIL$_{sf}$ & DIL$_{pf}$ & DIL$_{ss}$ & DIL$_{ps}$                       \\ 
\hline
\multirow{3}{*}{CBSD68~\cite{BSDall}}    
                          & 30  (\textit{unseen})                  & 24.90/0.581 & \textbf{30.29\textcolor{red}{$_{(5.39\uparrow)}$}/0.866}    & 29.92\textcolor{red}{$_{(5.02\uparrow)}$}/0.858 & 27.48\textcolor{red}{$_{(2.58\uparrow)}$}/0.809 & 29.14\textcolor{red}{$_{(4.24\uparrow)}$}/0.802                         \\
                          & 40   (\textit{unseen})                  & 21.12/0.400 & \textbf{28.35\textcolor{red}{$_{(7.23\uparrow)}$}/0.825} & 28.10\textcolor{red}{$_{(6.98\uparrow)}$}/0.812 & 25.90\textcolor{red}{$_{(4.78\uparrow)}$}/0.746 & 25.74\textcolor{red}{$_{(4.62\uparrow)}$}/0.629                        \\
                          & 50   (\textit{unseen})                  & 18.96/0.307 &        \textbf{26.64\textcolor{red}{$_{(7.68\uparrow)}$}/0.779}  & 26.61\textcolor{red}{$_{(7.65\uparrow)}$}/0.766 & 24.63\textcolor{red}{$_{(5.67\uparrow)}$}/0.686 & 23.34\textcolor{red}{$_{(4.38\uparrow)}$}/0.501                         \\ 
\hline
\multirow{3}{*}{Kodak24~\cite{Kodak}}
                          & 30  (\textit{unseen})                   & 25.12/0.533 & \textbf{31.39\textcolor{red}{$_{(6.27\uparrow)}$}/0.867} & 30.87\textcolor{red}{$_{(5.75\uparrow)}$}/0.858 & 27.92\textcolor{red}{$_{(2.80\uparrow)}$}/0.801 & 29.86\textcolor{red}{$_{(4.74\uparrow)}$}/0.782                        \\
                          & 40  (\textit{unseen})                   & 21.22/0.352 & \textbf{29.49\textcolor{red}{$_{(8.27\uparrow)}$}/0.831} & 29.15\textcolor{red}{$_{(7.93\uparrow)}$}/0.817 & 26.46\textcolor{red}{$_{(5.24\uparrow)}$}/0.738 & 26.13\textcolor{red}{$_{(4.91\uparrow)}$}/0.588                        \\
                           & 50  (\textit{unseen})                  & 19.02/0.263 & \textbf{27.76\textcolor{red}{$_{(8.74\uparrow)}$}/0.788} & 27.67\textcolor{red}{$_{(8.65\uparrow)}$}/0.775 & 25.24\textcolor{red}{$_{(6.22\uparrow)}$}/0.677 & 23.60\textcolor{red}{$_{(4.58\uparrow)}$}/0.457                         \\ 
\hline
\multirow{3}{*}{McMaster~\cite{McMaster}}
                          & 30  (\textit{unseen})                   & 25.65/0.569 & \textbf{31.70\textcolor{red}{$_{(6.05\uparrow)}$}/0.873} & 31.04\textcolor{red}{$_{(5.39\uparrow)}$}/0.853 & 28.15\textcolor{red}{$_{(2.50\uparrow)}$}/0.794 & 30.09\textcolor{red}{$_{(4.44\uparrow)}$}/0.800                         \\
                          & 40  (\textit{unseen})                   & 21.73/0.373 & \textbf{29.81\textcolor{red}{$_{(8.08\uparrow)}$}/0.831} & 29.07\textcolor{red}{$_{(7.34\uparrow)}$}/0.802 & 26.59\textcolor{red}{$_{(4.86\uparrow)}$}/0.728 & 26.24\textcolor{red}{$_{(4.51\uparrow)}$}/0.605                        \\
                          & 50   (\textit{unseen})                  & 19.47/0.278 & \textbf{28.02\textcolor{red}{$_{(8.55\uparrow)}$}/0.783} & 27.31\textcolor{red}{$_{(7.84\uparrow)}$}/0.749 & 25.20\textcolor{red}{$_{(5.73\uparrow)}$}/0.664 & 23.60\textcolor{red}{$_{(4.13\uparrow)}$}/0.466                         \\                         
\hline
\multirow{3}{*}{Urban100~\cite{urban100}} 
                          & 30  (\textit{unseen})                   & 25.46/0.648 & \textbf{30.93\textcolor{red}{$_{(5.47\uparrow)}$}/0.898} & 30.26\textcolor{red}{$_{(4.80\uparrow)}$}/0.884 & 26.95\textcolor{red}{$_{(1.49\uparrow)}$}/0.825 & 29.73\textcolor{red}{$_{(4.27\uparrow)}$}/0.841                         \\
                          & 40  (\textit{unseen})                   & 21.53/0.479 & \textbf{28.82\textcolor{red}{$_{(7.29\uparrow)}$}/0.866} & 28.32\textcolor{red}{$_{(6.79\uparrow)}$}/0.848 & 25.26\textcolor{red}{$_{(3.73\uparrow)}$}/0.767 & 26.25\textcolor{red}{$_{(4.72\uparrow)}$}/0.691                        \\
                          & 50  (\textit{unseen})                   & 19.28/0.389 & \textbf{26.88\textcolor{red}{$_{(7.60\uparrow)}$}/0.829} & 26.63\textcolor{red}{$_{(7.35\uparrow)}$}/0.811 & 23.85\textcolor{red}{$_{(4.57\uparrow)}$}/0.710 & 23.71\textcolor{red}{$_{(4.43\uparrow)}$}/0.575                         \\ 
\hline
\multirow{3}{*}{Manga109~\cite{manga109}} 
                          & 30  (\textit{unseen})                   & 26.62/0.653 & \textbf{31.97\textcolor{red}{$_{(5.35\uparrow)}$}/0.910} & 31.14\textcolor{red}{$_{(4.52\uparrow)}$}/0.901 & 26.02\textcolor{red}{$_{(-0.6\uparrow)}$}/0.833 & 31.05\textcolor{red}{$_{(4.43\uparrow)}$}/0.858                         \\
                          & 40  (\textit{unseen})                   & 22.34/0.442 & \textbf{29.02\textcolor{red}{$_{(6.68\uparrow)}$}/0.888} & 28.53\textcolor{red}{$_{(6.19\uparrow)}$}/0.875 & 24.31\textcolor{red}{$_{(1.97\uparrow)}$}/0.784 & 27.29\textcolor{red}{$_{(4.95\uparrow)}$}/0.704                        \\
                          & 50  (\textit{unseen})                   & 19.95/0.342 & \textbf{26.52\textcolor{red}{$_{(6.57\uparrow)}$}/0.860} & 26.34\textcolor{red}{$_{(6.39\uparrow)}$}/0.846 & 22.82\textcolor{red}{$_{(2.87\uparrow)}$}/0.734 & 24.47\textcolor{red}{$_{(4.52\uparrow)}$}/0.564                         \\ 
\hline
\end{tabular}}
\label{table:noise}
\vspace{-4mm}
\end{table*}

\subsection{Implementations of DIL from Meta-Learning}
An interesting finding is that the derived optimization direction of DIL from causality perspective in Eq.~\ref{equ:final-update} is consistent with one typical meta-learning strategy termed as MAML~\cite{finn2017modelMAML}, \tcb{even they have} different purposes. MAML aims to enable the fast adaptation capability of a network for few-shot tasks, while ours aims to improve the generalization capability of the restoration network. We \tcb{facilitate} our DIL in image restoration \tcb{based on} this meta-learning strategy.

However, it is arduous to directly incorporate the optimization direction of Eq.~\ref{equ:final-update} into the practical training process, which is computationally prohibitive. The reason is that it requires multiple gradient computing and updating, which is expensive, especially for the pixel-wise image restoration. To simplify this process, we utilize the Talyor expansion and inverse expansion to derive Eq.~\ref{equ:final-update} as:
\begin{figure}[htp]
    \centering
    \includegraphics[width=0.90\linewidth]{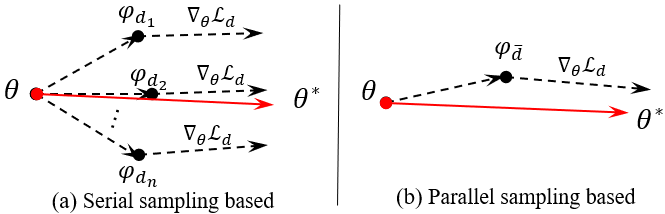}
    \caption{The comparison of serial sampling and parallel sampling.}
    \label{fig:sampling}
    \vspace{-8mm}
\end{figure}
\begin{equation}
 \begin{aligned}
      &\theta^* = \arg\min_{\theta}  \mathbb{E}_{(I_d, I_c)\sim \mathcal{D}} [  \mathcal{L}(f_{\phi_{\Bar{d}}}(I_d), I_c)], \\ 
       & where ~\phi_{\Tilde{d}}= \theta - \alpha \nabla_\theta \sum_{d_i\in D}\frac{1}{n}\mathcal{L}(f_\theta(I_{d_i}), I_c),
\end{aligned}  
\label{eq:simplify}
\end{equation}
where $\phi_{\Bar{d}}$ denotes the parameters of restoration network $f$ that is virtually updated with loss function with samples overall all distortions $D=\{d_i\}, 1\le i \le n$. We define it as \textit{parallel sampling} for DIL, which reduces the complex training process of \textbf{DIL} to two steps. In this paper, we call the original sampling strategy as \textit{serial sampling}. The comparison between \textit{serial sampling} and \textit{parallel sampling} are shown in Fig.~\ref{fig:sampling}. The detailed derivation for Eq.~\ref{eq:simplify} are described in the \textbf{Supplementary}. 

\tcr{We also investigate two different gradient updating strategy for DIL. From Eq.~\ref{equ:final-update} and Eq.~\ref{eq:simplify}, we can observe that they require the second-order gradient since the gradient is computed with two-step forward through $\phi_{d_i}$, which is shown in Fig.~\ref{fig:sampling}. To simplify it, Reptile~\cite{nichol2018firstReptile} proposes an alternative strategy (\textit{i.e.,} approximating the second-order gradient by the sequential parameter updating with one-order gradient. The optimization direction (\textit{i.e.}, gradient) is computed with the deviation between the initial and last-step parameters. We integrate it into our DIL and call it first-order optimization. In contrast, the original optimization in Eq.~\ref{eq:simplify} is termed second-order optimization.}
In summary, we propose four variants for DIL following the above two strategies.  DIL$_{sf}$ adopts the \textbf{s}erial sampling and \textbf{f}irst-order gradient optimization. DIL$_{pf}$ utilizes the \textbf{p}arallel sampling and \textbf{f}irst-order optimization. DIL$_{ss}$/DIL$_{ps}$ exploits the \textbf{s}econd-order optimization and \textbf{s}erial/\textbf{p}arallel sampling. 

\begin{table*}[ht]
\centering
\caption{Quantitative comparison for image deblurring on several benchmark datasets. Results are tested on the five unseen blur degrees [4.2, 4.4, 4.6, 4.8, 5.0] in terms of PSNR/SSIM on RGB channel.}
\resizebox{0.95\textwidth}{!}{\begin{tabular}{c|c|c|c|c|c|c} 
\hline
\multirow{2}{*}{Datasets} & \multirow{2}{*}{Methods} & \multicolumn{5}{c}{Levels}                                \\ 
\cline{3-7} &   & 4.2 (\textit{unseen}) & 4.4 (\textit{unseen}) & 4.6 (\textit{unseen}) & 4.8 (\textit{unseen}) & 5.0 (\textit{unseen}) \\ 
\hline
\multirow{2}{*}{Set5~\cite{Set5}}    
                          & ERM                 & 29.31/0.844 & 26.55/0.776    & 24.43/0.709 & 22.96/0.648 & 22.00/0.602                         \\ 
                          & \cellcolor{gray!20}  \textbf{DIL}                 & \cellcolor{gray!20} 29.58\textcolor{red}{$_{(0.27\uparrow)}$}/0.848 & \cellcolor{gray!20} 27.52\textcolor{red}{$_{(0.97\uparrow)}$}/0.802 & \cellcolor{gray!20} 25.66\textcolor{red}{$_{(1.23\uparrow)}$}/0.751 & \cellcolor{gray!20} 24.38\textcolor{red}{$_{(1.42\uparrow)}$}/0.708 & \cellcolor{gray!20} 23.46\textcolor{red}{$_{(1.46\uparrow)}$}/0.671                        \\

\hline
\multirow{2}{*}{Set14~\cite{Set14}}
                          & ERM                   & 27.22/0.781 & 24.93/0.726 & 23.16/0.671 & 21.89/0.624 & 20.88/0.583                        \\
                          & \cellcolor{gray!20}  \textbf{DIL}                   & \cellcolor{gray!20} 27.24\textcolor{red}{$_{(0.02\uparrow)}$}/0.778 & \cellcolor{gray!20} 25.78\textcolor{red}{$_{(0.85\uparrow)}$}/0.746 & \cellcolor{gray!20} 24.35\textcolor{red}{$_{(1.19\uparrow)}$}/0.708 & \cellcolor{gray!20} 23.23\textcolor{red}{$_{(1.34\uparrow)}$}/0.672 & \cellcolor{gray!20} 22.37\textcolor{red}{$_{(1.49\uparrow)}$}/0.640                        \\
\hline
\multirow{2}{*}{BSD100~\cite{BSDall}}
                          & ERM                   & 27.20/0.784 & 25.17/0.732 & 23.50/0.682 & 22.24/0.639 & 21.28/0.602                         \\
                          & \cellcolor{gray!20}  \textbf{DIL}                   & \cellcolor{gray!20} 27.37\textcolor{red}{$_{(0.17\uparrow)}$}/0.781 & \cellcolor{gray!20} 26.16\textcolor{red}{$_{(0.99\uparrow)}$}/0.753 & \cellcolor{gray!20} 24.91\textcolor{red}{$_{(1.41\uparrow)}$}/0.719 & \cellcolor{gray!20} 23.86\textcolor{red}{$_{(1.62\uparrow)}$}/0.686 & \cellcolor{gray!20} 23.02\textcolor{red}{$_{(1.74\uparrow)}$}/0.658                        \\
\hline
\multirow{2}{*}{Urban100~\cite{urban100}} 
                          & ERM                   & 24.95/0.797 & 22.41/0.723 & 20.59/0.657 & 19.33/0.606 & 18.40/0.565                         \\
                          &\cellcolor{gray!20} \textbf{DIL}                   & \cellcolor{gray!20} 24.97\textcolor{red}{$_{(0.02\uparrow)}$}/0.793 & \cellcolor{gray!20} 23.26\textcolor{red}{$_{(0.85\uparrow)}$}/0.743 & \cellcolor{gray!20} 21.76\textcolor{red}{$_{(1.17\uparrow)}$}/0.693 & \cellcolor{gray!20} 20.70\textcolor{red}{$_{(1.37\uparrow)}$}/0.651 & \cellcolor{gray!20} 19.92\textcolor{red}{$_{(1.52\uparrow)}$}/0.618                        \\
\hline
\multirow{2}{*}{Manga109~\cite{manga109}} 
                          & ERM                   & 28.16/0.865 & 23.96/0.791 & 21.21/0.713 & 19.63/0.652 & 18.63/0.606                         \\
                          & \cellcolor{gray!20} \textbf{DIL}                   & \cellcolor{gray!20} 28.09\textcolor{red}{$_{(-0.07\uparrow)}$}/0.867 & \cellcolor{gray!20} 25.41\textcolor{red}{$_{(1.45\uparrow)}$}/0.822 & \cellcolor{gray!20} 23.15\textcolor{red}{$_{(1.94\uparrow)}$}/0.771 & \cellcolor{gray!20} 21.69\textcolor{red}{$_{(2.06\uparrow)}$}/0.726 & \cellcolor{gray!20} 20.72\textcolor{red}{$_{(2.09\uparrow)}$}/0.691     \\
\hline 
\end{tabular}}
\label{table:blur}
\vspace{-3mm}
\end{table*}
\begin{figure*}[ht]
    \centering
\includegraphics[width=0.95\linewidth]{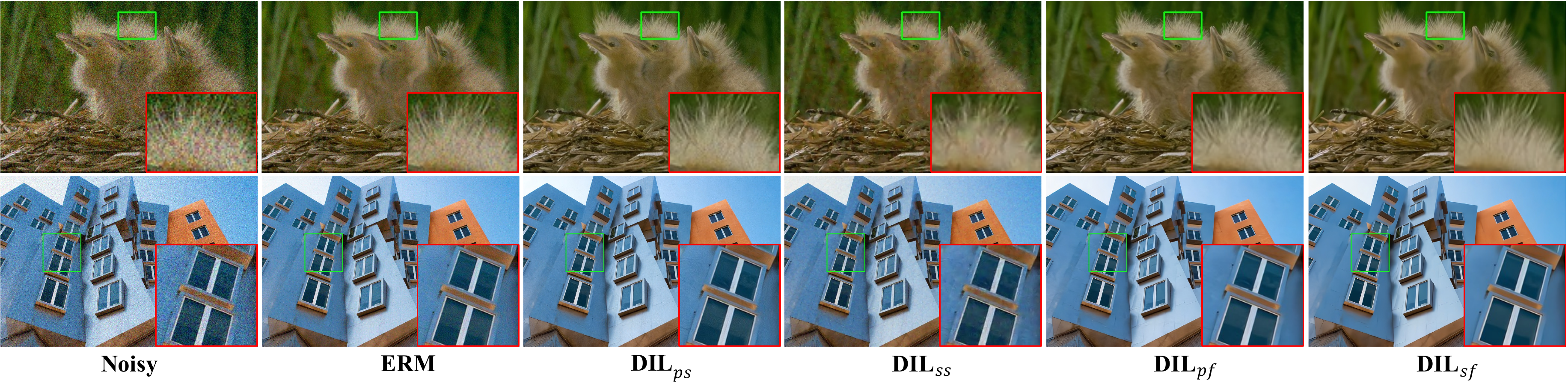}
    \caption{Visual comparison of the commonly-used ERM and our proposed four variants of DIL with the unseen noise level 30.}
    \label{fig:noise}
    \vspace{-5mm}
\end{figure*}
\section{Experiments}
In this section. we first describe the implementation details. Then, we validate the effectiveness of our DIL from two typical out-of-distribution settings, \textit{i.e.,} Cross Distortion Degrees, and Cross Distortion Types. Particularly, for cross-distortion degrees, we train the restoration network with seen distortion degrees while testing it with unseen distortion degrees. For cross-distortion types, the restoration network is trained with synthesized distortions and validated on the corresponding real-world or other distortions.

\subsection{Implementation}
 We adopt the typical RRDB~\cite{ESRGAN} as our image restoration backbone, which has demonstrated remarkable performances towards various low-level image tasks~\cite{RealESRGAN,wang2019deformable}. All the experiments are done with four NVIDIA 2080Ti GPUs. 
 Adam optimizer is adopted to optimize network parameters in both ERM and \textbf{DIL} training paradigms. More details are given in the \textbf{Supplementary}.

\subsection{Cross Distortion Degrees}
\noindent\textbf{Results on Image Denoising.}
For image denoising, the training data are composed of distorted images with noise levels [5, 10, 15, 20] and their corresponding clean images. After training the restoration network, we validate it on the test datasets with unseen noise degrees, including [30, 40, 50]. We compare the empirical risk minimization (ERM) and four variants of our proposed DIL, \textit{i.e.,} DIL$_{sf}$, DIL$_{pf}$, DIL$_{ss}$, and DIL$_{ps}$, respectively. 

\begin{table}[h!]
\centering
\caption{Quantitative comparison for hybrid distortion removal. Results are tested on three different distortion levels in terms of PSNR/SSIM on Y channel. }
\resizebox{\linewidth}{!}{\setlength{\tabcolsep}{0.5mm}{\begin{tabular}{c|c|c|c|l} 
\hline
\multirow{3}{*}{Datasets} & \multirow{3}{*}{Methods} & \multicolumn{3}{c}{Distortion level}            \\ 
\cline{3-5}
                          &                          & \makecell{Mild \\ (unseen)} & \makecell{Moderate\\(unseen)} & \makecell{Severe \\ (seen)} \\ 

\hline
\multirow{2}{*}{BSD100~\cite{BSDall}}   & ERM                      & 25.31/0.687  & 24.62/0.642 & 25.27/0.617 \\
                          & 
                          \cellcolor{gray!20} \textbf{DIL}                      &\cellcolor{gray!20}26.37/0.691 &\cellcolor{gray!20}25.23/0.645 & \cellcolor{gray!20}25.22/0.613 \\ 
\hline
\multirow{2}{*}{Urban100~\cite{urban100}} & ERM                      & 23.97/0.736 & 22.51/0.674 & 23.38/0.655 \\
                          &\cellcolor{gray!20} \textbf{DIL}                      & \cellcolor{gray!20}25.00/0.747 &\cellcolor{gray!20}23.13/0.682 & \cellcolor{gray!20}23.20/0.645 \\ 
\hline
\multirow{2}{*}{Manga109~\cite{manga109}} & ERM                      & 27.43/0.863 & 24.85/0.808 & 26.50/0.815 \\
                          &\cellcolor{gray!20}  \textbf{DIL}                      & \cellcolor{gray!20}28.41/0.868 & \cellcolor{gray!20}25.30/0.810 & \cellcolor{gray!20}26.19/0.766 \\
\hline
\multirow{2}{*}{DIV2K~\cite{DIV2K}}   & ERM                      & 26.19/0.766 & 25.94/0.744 & 27.42/0.742 \\
                          & \cellcolor{gray!20} \textbf{DIL}                      &\cellcolor{gray!20}27.84/0.785 &\cellcolor{gray!20}26.89/0.756 &\cellcolor{gray!20}27.38/0.737 \\
\hline
\end{tabular}}}
\label{table:hybrid}
\vspace{-3mm}
\end{table}

\begin{table*}[h]
\caption{Quantitative results of network generalization capability on real image denoising and synthetic image deraining tasks. Results are tested on Y channel in terms of PSNR/SSIM, except for DND where we obtain our results from official online benchmark.}
\centering
\resizebox{0.8\textwidth}{!}{\setlength{\tabcolsep}{1.0mm}{\begin{tabular}{c|c|c|c|c|c|c} 
\hline
\multirow{2}{*}{Methods} & \multicolumn{2}{c|}{Datasets (Real Denoising)}  & & \multicolumn{3}{c}{Datasets (Deraining)}                             \\ 
\cline{2-3} \cline{5-7}
                          & SIDD~\cite{SIDD}  & DND~\cite{DND} & & Rain100L~\cite{Rain100} & Rain12~\cite{Rain12} & Rain800~\cite{Rain800} \\ 
\cline{1-3} \cline{5-7}
ERM           & 38.90/0.9379 & 38.67/0.9549 && 27.61/0.8577 & 31.44/0.8947 & 23.36/0.8199  \\
\cline{1-3} \cline{5-7}
DIL$_{sf}$           & 39.96\textcolor{red}{$_{(1.06\uparrow)}$}/0.9410 & 39.16\textcolor{red}{$_{(0.49\uparrow)}$}/0.9531 & & 28.15\textcolor{red}{$_{(0.54\uparrow)}$}/0.8679  & 32.43\textcolor{red}{$_{(0.99\uparrow)}$}/0.9163 & 23.41\textcolor{red}{$_{(0.05\uparrow)}$}/0.8261 \\
\rowcolor{gray!20}DIL$_{ps}$           & 39.92\textcolor{red}{$_{(1.02\uparrow)}$}/0.9385 & 39.03\textcolor{red}{$_{(0.36\uparrow)}$}/0.9553 & &  28.37\textcolor{red}{$_{(0.76\uparrow)}$}/0.8739 & 33.07\textcolor{red}{$_{(1.63\uparrow)}$}/0.9266 & 23.52\textcolor{red}{$_{(0.16\uparrow)}$}/0.8281 \\
\hline
\end{tabular}}}
\label{table:realnoise}
\end{table*}
\begin{figure*}[ht]
    \centering
    \includegraphics[width=0.8\linewidth]{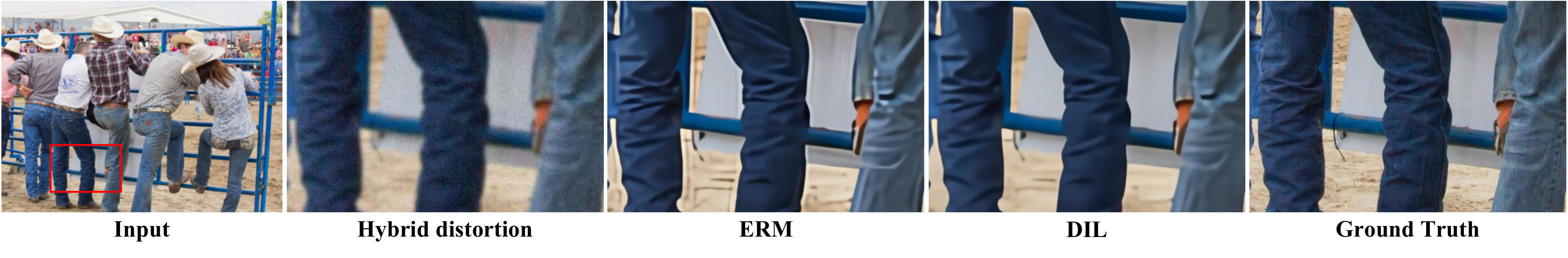}
    \caption{Visual comparison of the commonly-used ERM and our proposed DIL for unseen hybrid-distorted (mild) image restoration.}
    \label{fig:hybrid}
    \vspace{-6mm}
\end{figure*}
The experimental results are shown in Table~\ref{table:noise}. We can observe that all four variants of DIL achieve great generalization ability on multiple unseen noise levels compared with commonly-used empirical risk minimization (ERM). On several typical scenarios, including natural images (\textit{i.e.,} CBSD68~\cite{BSDall}, Kodak24~\cite{Kodak}, McMaster~\cite{McMaster}), building images (Urban100~\cite{urban100}), cartoon images (\textit{i.e.,} Manga109~\cite{manga109}), our DIL even outperforms the ERM by a promising/amazing gain of \tcr{8.74} dB at most. Moreover, with the increase of the distribution gap between training and testing data, ours can achieve larger improvements for ERM. Furthermore, for cross distortion degree, DIL$_{sf}$ shows the best generalization capability compared with the other three variants by serial sampling and first-order optimization. We also visualize the reconstructed images of the above methods in Fig.~\ref{fig:noise}.
For the unseen distortion degree ($\sigma=30$), the ERM cannot remove the noise well and the reconstructed image also contains obvious noise distortion. However, our DIL$_{sf}$ enables the restoration network to recover more vivid and clean images from the unseen noise degrees, which validates the correctness and effectiveness of our proposed DIL. 

\noindent\textbf{Results on Image Deblurring.}
\label{sec:deburring}
We also validate the generalization capability of our DIL on the challenging image deblurring. 
Under this scenarios, we train the restoration network with our proposed DIL with the gaussian blurring level [1.0, 2.0, 3.0, 4.0], and validate its generalization capability on the more severe and difficult blurring levels, including 4.2, 4.4, 4.6, 4.8, and 5.0. 

As shown in Table.~\ref{table:blur}, we validate our DIL on five benchmark datasets, including Set5~\cite{Set5}, Set14~\cite{Set14}, BSD100~\cite{BSDall}, Urban100~\cite{urban100}, and Manga109~\cite{manga109}.  With the increase of blurring level, the restoration network trained with ERM suffers from a severe performance drop, since the unseen blurring levels are far away from the blurring levels used for training. But our DIL can improve ERM on each unseen blurring level for five datasets. In particular, we achieve the gain of 2.09 dB for the cartoon scene Manga109~\cite{manga109} on the blurring level 5.0. 


\noindent\textbf{Results on Hybrid-distorted Image Restoration.}
Except for the above single distortion, we also explore the generalization capability of our DIL on hybrid-distorted image restoration. Following~\cite{li2020learningFDRNet}, 
the hybrid distorted images are degraded with  blur, noise, and Jpeg compression in a sequence manner. Based on the distortion degree, it can be divided into three levels from low to high, \textit{i.e.,} mild, moderate, and severe. In this setting, the restoration network is trained with severe hybrid distortions and validated on the mild and moderate levels.

As shown in Table~\ref{table:hybrid}, our DIL achieves an average gain of 1.05 dB, and 0.66 dB on the mild-level, and moderate-level hybrid distortions than ERM, which has a large distribution gap with severe-level hybrid distortions. We can also notice that with the increase of the distribution gap, ours can preserve more performances on the restoration of the out-of-distribution distortions. We also conduct the subjective comparison of our methods with the commonly-used ERM in Fig.~\ref{fig:hybrid}. We can observe that the restoration network trained with ERM suffers from new artifacts for unseen hybrid-distorted images. But our DIL can eliminate the artifacts well and generate more promising results.

\begin{figure*}[ht]
    \centering
    \includegraphics[width=0.98\linewidth]{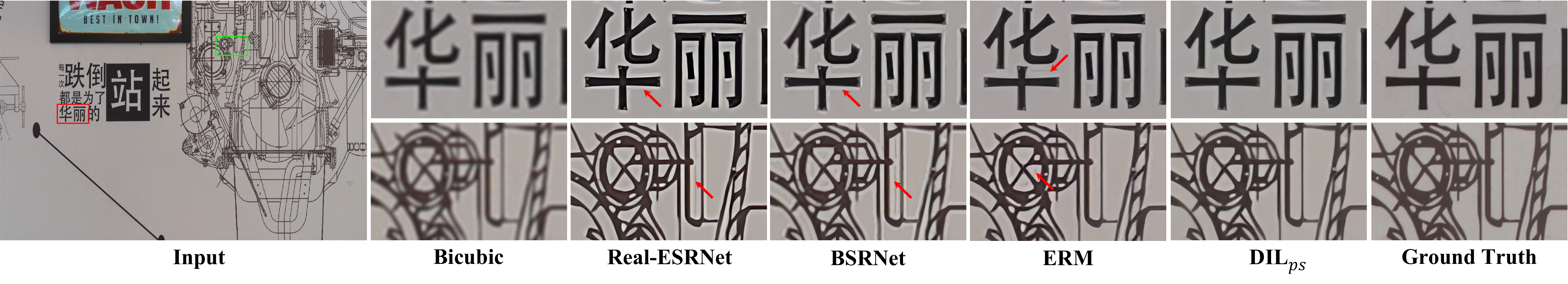}
    \caption{Visual comparison with state-of-the-art methods on DRealSR~\cite{DrealSR}.}
    \label{fig:realsr}
    \vspace{-6mm}
\end{figure*}

\subsection{Cross Distortion Types}
In this section, we investigate the effects of our proposed DIL on the cross-distortion type setting, which is more challenging than the cross-degree setting. 

\noindent\textbf{Results on Real Image Super-resolution}
Real Image Super-resolution (RealSR) has attracted 
great attention since it is urgently required in real life, where the distorted image contains complex hybrid distortions, such as blurring, low resolution, noise, etc. However, the distorted/clean pairs for RealSR are hard to be collected. Simulating distortions like Real-world distortion has been a popular solution for RealSR~\cite{RealESRGAN,BSRGAN}. In this paper, we follow the Real-ESRGAN~\cite{RealESRGAN} and utilize its proposed RealSR distortion simulating to generate image pairs as training datasets. Then we test the restoration network on the out-of-distribution datasets, RealSR V3~\cite{RealSR}, DRealSR~\cite{DrealSR}, which are two commonly-used datasets for RealSR evaluation.

\begin{table}[h]
\caption{Quantitative results of the network generalization capability on RealSR tasks. Results are tested on the Y channel in terms of PSNR/SSIM.}
\centering
\resizebox{0.95\linewidth}{!}{\setlength{\tabcolsep}{0.5mm}{\begin{tabular}{c|c|c} 
\hline
\multirow{2}{*}{Methods} & \multicolumn{2}{c}{Datasets}                                \\ 
\cline{2-3}
                          & RealSR V3~\cite{RealSR} (unseen)  & DrealSR~\cite{DrealSR} (unseen) \\ 
\hline
Real-ESRNet~\cite{RealESRGAN}               & 26.19/0.7989 & 28.22/0.8470  \\
BSRNet~\cite{BSRGAN}           & 27.46/0.8082 & 29.45/0.8579 \\
\hline
ERM           & 27.65/0.8098 & 29.73/0.8628 \\
DIL$_{sf}$           & 27.94\textcolor{red}{$_{(0.29\uparrow)}$}/0.8098 & 29.99\textcolor{red}{$_{(0.26\uparrow)}$}/0.8648 \\

\rowcolor{gray!20} DIL$_{ps}$           & 28.12\textcolor{red}{$_{(0.47\uparrow)}$}/0.8067 & 30.58\textcolor{red}{$_{(0.85\uparrow)}$}/0.8712\\
\hline
\end{tabular}}}
\label{table:realsr}
\vspace{-5mm}
\end{table}


\begin{table}[h]
\caption{Quantitative results of our DIL on different backbones. Results are tested on the unseen noise level 30 in terms of PSNR/SSIM.}
\centering
\resizebox{0.9\linewidth}{!}{\setlength{\tabcolsep}{0.5mm}{\begin{tabular}{c|c|c|c|c} 
\hline
\multirow{2}{*}{Models} & \multirow{2}{*}{Methods} & \multicolumn{3}{c}{Datasets}                                \\ 
\cline{3-5}
                          & & CBSD68~\cite{BSDall}  & Kodak24~\cite{Kodak} & Urban100~\cite{urban100} \\ 
\hline
\multirow{2}{*}{RRDB}              & ERM & 24.90/0.581 & 25.12/0.533 & 25.46/0.648  \\
&\cellcolor{gray!20}  \textbf{DIL} & \cellcolor{gray!20}30.28/0.866 &\cellcolor{gray!20}31.39/0.867 &\cellcolor{gray!20}30.93/0.898 \\
\hline
\multirow{2}{*}{SwinIR}          & ERM & 24.22/0.551 & 24.22/0.493 & 24.73/0.618\\
&\cellcolor{gray!20}  \textbf{DIL} &\cellcolor{gray!20}29.08/0.798 &\cellcolor{gray!20}29.71/0.774 & \cellcolor{gray!20}29.72/0.834\\
\hline
\end{tabular}}}
\label{table:backbone}
\vspace{-4 mm}
\end{table}

We show the experimental results on RealSR in Table.~\ref{table:realsr}. Without access to any training samples in RealSR V3, DRealSR, our DIL$_{sf}$ can outperform the ERM by 0.29dB on RealSR V3~\cite{RealSR} and 0.26dB on DRealSR dataset~\cite{DrealSR}. Particularly, we  notice that DIL$_{ps}$ is more suitable for cross-distortion type scenarios than DIL$_{sf}$, which exceeds the ERM by a 0.47dB on RealSR V3, and 0.85dB on DRealSR dataset. 
The reason for that we guess is that  DIL$_{ps}$ is more capable of improving the generalization for the large distribution gap in image restoration. We also visualize the comparison corresponding to the subjective quality for different methods. As shown in Fig.~\ref{fig:realsr}, Real-ESRNet~\cite{RealESRGAN} and BSRNet~\cite{BSRGAN} cause the overshooting at the edge 
of the text. But our DIL$_{ps}$ can eliminate the artifacts and achieve a high-quality restoration
\begin{table}[h]
\caption{Quantitative comparison between different distortion augmentation methods. D$_1$ and D$_2$ are the first order distortion and the second order distortion derived from~\cite{RealESRGAN} respectively. Results are tested on RealSR datasets in terms of PSNR/SSIM.}
\centering
\resizebox{0.85\linewidth}{!}
{\begin{tabular}{c|c|c|c} 
\hline
\multirow{2}{*}{Augmentation} & \multirow{2}{*}{Methods} & \multicolumn{2}{c}{Datasets}                                \\ 
\cline{3-4}
                          & & RealSR V3~\cite{RealSR}  & DrealSR~\cite{DrealSR} \\ 
\hline
\multirow{2}{*}{D$_1$}              & ERM & 27.65/0.8098 & 29.73/0.8628  \\
& \cellcolor{gray!20} \textbf{DIL} & \cellcolor{gray!20}27.94/0.8098 & \cellcolor{gray!20}29.99/0.8648 \\\hline
\multirow{2}{*}{D$_2$}          & ERM & 27.39/0.8077 & 29.41/0.8591 \\
& \cellcolor{gray!20} \textbf{DIL} &\cellcolor{gray!20}27.65/0.8027 &\cellcolor{gray!20}29.85/0.8677 \\
\hline
\end{tabular}}
\label{table:aug}
\vspace{-4mm}
\end{table}

\noindent\textbf{Results on Real Image Denoising.}
We also study the generalization capability of our training paradigm DIL on the Real Image Denoising task. Concretely, we select four synthesized distortions based on four categories of color space among camera ISP process~\cite{CBDNet}, and generate training image pairs from DF2K~\cite{DIV2K,Flickr2K} in an online manner. Then we verify its generalization on the commonly-used Real Denoising dataset SIDD~\cite{SIDD} and DND~\cite{DND}.  
As Table~\ref{table:realnoise} illustrated, our DIL$_{ps}$ achieves the PSNR of 39.92 dB, which outperforms the ERM by 1.02dB, which is almost the same with DIL$_{sf}$.

\noindent\textbf{Results on Image Deraining.}
As an extension experiment, we introduce our DIL to the experiments of
image deraining task. Particularly, the raining types and degrees between different datasets are severely different in image deraining. Here, we optimize the restoration network with three image deraining datasets, including DID-MDN~\cite{DID-MDN}, Rain14000~\cite{Rain12600}, and Heavy Rain Dataset~\cite{heavyrain}. Then we validate the generalization capability of the restoration network on three unseen deraining datasets, \textit{i.e.,} Rain100L~\cite{Rain100}, Rain12~\cite{Rain12}, and Rain800~\cite{Rain800}. We report the experimental results in Table~\ref{table:realnoise}. Our DIL (DIL$_{ps}$) enables the restoration network to have a better generalization capability than ERM, which obtains a gain of 0.76dB on Rain100L~\cite{Rain100} and 1.63dB on Rain12~\cite{Rain12} dataset.


\subsection{Ablation Studies}
\noindent\textbf{Impact of different restoration networks.}
We demonstrate the effectiveness of DIL across different network backbones. In addition to the convolution-based RRDB~\cite{ESRGAN} network, we also incorporate our DIL into the transformer-based SwinIR~\cite{liang2021swinir}. The performances are reported in Table~\ref{table:backbone}, which reveals that our DIL can also improve the generalization capability of Transformer-based backbones. This study reveals our DIL is a general training paradigm for different backbones. 

\noindent\textbf{Effects of different variants for DIL}
As shown in Table.~\ref{table:noise},and~\ref{table:realnoise}, we can observe that DIL$_{sf}$ is more proper for cross-distortion degrees. But for cross-distortion types, DIL$_{ps}$ achieves better performance for RealSR and Image Deraining. It is noteworthy that the distribution gap of different distortion types is larger than different degrees. The first-order optimization is more stable but lacks enough capability for a severe distribution gap compared to second-order optimization. But all of them are competent in improving the generalization capability.  


\vspace{-3mm}
\section{Discussion on Limitations}
\vspace{-2mm}
\noindent\textbf{The performance on training data.}
We also report the performance of our DIL on the seen training data in Table~\ref{table:hybrid}. It can be seen that our DIL will cause a slight performance drop but the generalization capability is improved obviously. The reason for that is our DIL implements distortion invariant representation learning, which prevents the restoration network from over-fitting to the training data.

\noindent\textbf{The impact of different distortion augmentation.} As shown in Table~\ref{table:aug}, despite that our DIL achieves the improvement of the generalization capability. The final generalization performance is still related to the distortion augmentation strategy. It is vital to find a universal distortion augmentation strategy, which requires more exploration. We believe it will be a potential/important direction to improve the generalization ability of the restoration network. 

\vspace{-3mm}
\section{Conclusion}
\vspace{-2mm}
In this paper, we propose a novel \textit{distortion invariant representation learning} (\textbf{DIL}) training paradigm for image restoration from the causality perspective. In particular, we provide a causal view of the image restoration process, and clarify why the restoration network lacks the generalization capability for different degradations. Based on that, we treat the distortion types and degrees as confounders, of which the confounding effects can be removed with our proposed \textbf{DIL}. Concretely, we produce the spurious confounders by simulating the different distortion types and degrees. Then, an instantiation of the back-door criterion in causality is introduced from the  optimization perspective, which
enables the restoration network to remove the harmful bias from different degradations. 
Extensive experiments on the settings, cross distortion degrees, and cross distortion types, have demonstrated that our \textbf{DIL} improves the generalization capability of the restoration network effectively.   

\section*{Acknowledgements}
This work was supported in part by NSFC under Grant U1908209, 62021001, and  ZJNSFC under Grant LQ23F010008.

{\small
\bibliographystyle{ieee_fullname}
\bibliography{egbib}
}

\clearpage
\section*{Appendix}

\noindent Section~\ref{sec:back-door} provides the systematic introduction for the related notations of the back-door criterion in causal learning.

\noindent Section~\ref{sec:proof} explains the counterfactual distortion augmentation from the causality perspective. 

\noindent Section~\ref{sec:derivation} theoretically derives the parallel sampling in Eq. 5 of our paper.

\noindent Section~\ref{sec:algs} clarifies the implementations of four variants of our DIL, which can help the readers to reproduce our methods more easily. 

\noindent Section~\ref{sec:implementation} describes the more detailed experimental settings and the construction of distortion/confounder set $D$ in different image restoration tasks.

\noindent Section~\ref{sec:subjective} visualizes more subjective comparisons on different image restoration tasks.

\section{The Back-door Criterion in Causality.}
\label{sec:back-door}
In this section, we clarify the related notations and derivations for the back-door criterion in causality.

\begin{figure}
    \centering
    \includegraphics[width=0.4\linewidth]{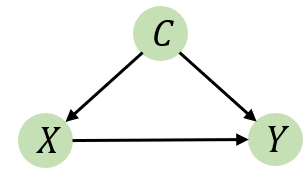}
    \caption{A structural causal model for back-door structure.}
    \label{fig:back_door}
\end{figure}

\noindent\textbf{Structure causal Model.}
As described in~\cite{pearl2009causal,glymour2016causality}, we can describe the causal relationship between different vectors with a directed Structural causal Model (SCM) like Fig.~\ref{fig:back_door}. A directed arrow $X\xrightarrow[]{}Y$ represent $X$ is the cause of the $Y$. The difference between correlation and causation is as follows: 
1) In causation, given $X\xrightarrow[]{}Y$, changing the $X$ will cause the effect on $Y$. But changing $Y$ does not have an effect on $X$ since $Y$ is not the cause of $X$. 
2) In correlation, we can compute the correlation between $X$ and $Y$ with conditional probability $P(Y|X)$ and $P(X|Y)$ no matter whether there is causation between $X$ and $Y$. In general, model training in deep learning is a process to fit the correlation between inputs and their labels instead of the causation. 

\noindent\textbf{Confounder.}
The confounder is defined based on the SCM, which represents the variables (\textit{e.g.}, $C$ in Fig.~\ref{fig:back_door}) that are the common cause between two other variables (\textit{e.g.,} $X$ and $Y$ in Fig.~\ref{fig:back_door}). The fork connection $X\xleftarrow[]{}C\xrightarrow[]{}Y$ causes the spurious correlation for $X$ and $Y$, which has a confounding effect on the estimation of the causal relationship between $X$ and $Y$. In other words, the correlation between $X$ and $Y$ learned by the model also is implicitly conditioned on the confounder $C$. 

\begin{figure}
    \centering
    \includegraphics[width=0.9\linewidth]{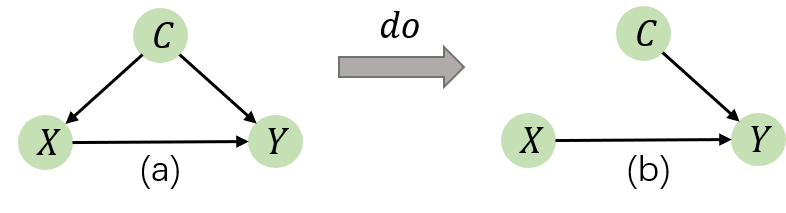}
    \caption{Back-door Criterion in Causality.}
    \label{fig:back_door_criterion}
\end{figure}

\noindent\textbf{\textit{do} operation.} A \textit{do} operation means to cut off the connection from the $C\xrightarrow[]{}X$, which is shown in Fig.~\ref{fig:back_door_criterion}. In this way, the correlation introduced from the path $X\xleftarrow[]{}C\xrightarrow[]{}Y$ is removed from \textit{do} operation. Then the correlation learned by the model is only from the $X\xrightarrow[]{}Y$, which are represented as $P(Y|do(X))$. And this causal correlation is independent of the confounder $C$ and is what we expect the model to learn. 
 
\noindent\textbf{Back-door criterion.}
The back-door criterion is proposed in~\cite{pearl2009causal,glymour2016causality}, which aims to implement the \textit{do} operation and eliminate the spurious correlation existed in $X\xleftarrow[]{}C\xrightarrow[]{}Y$. It removes the confounding effects of confounder $C$ by computing the average causal effects between $X\xrightarrow[]{}Y$ by traversing all values of $C$ as:
\begin{equation}
   P(Y|do(X)) = \sum_{c} P(Y|X,C=c)P(C=c)
    \label{eq:1}
\end{equation} 
Based on Eq.~\ref{eq:1}, we can achieve the \textit{do} operation in Fig.~\ref{fig:back_door_criterion} (b).

\noindent\textbf{The back-door criterion in Image Restoration}
As shown in Fig. 2 in our paper, we model the image restoration process as a structural causal model, where $D=\{d_i|1\le i \le n\}$ are the confounders between the distorted images $I_d$ and the expected reconstruction images $I_o$, which satisfies the back-door criterion. Therefore, we can derive the back-door criterion in image restoration as: 
\begin{equation}
   P(I_o|do(I_d)) = \sum_{i=1}^{n} P(I_o|I_d,D=d_i)P(D=d_i)
    \label{eq:2}
\end{equation}

\section{A proof for counterfactual distortion augmentation.}
\label{sec:proof}
The conterfactuals aims to answer the question "\textit{``if $X$ been $x$, in the situation $U$, what $Y_{X=x}(U)$ would be?"}. The three variables are in the same structural causal model (SCM), and $X$ and $U$ are the cause of $Y$. As described in ~\cite{pearl2009causal,glymour2016causality}, the calculating of counterfactuals follows three steps:
1) Abduction: Use evidence $e$ to determine the value of U.
2) Action: Remove the structural equations for the variables X to modify the model $M$ (\textit{i.e.,} the SCM). Then, set the $X$ as $X=x$ to obtain the modified $M_x$.
3) Prediction: Use the $M_x$ and $U=u$ to compute the value of $Y$ (\textit{i.e.,} the consequence of the counterfactual).

Considering the generation process of the distorted images $I_d = g(I_c, d)$, where $I_c$ and $d$ are the clean images and distortion type/degree, respectively. $g$ is the degradation process. The generation process can be modeled as a structural causal model $I_c\xrightarrow[]{}I_d\xleftarrow[]{}d$. To construct the datasets for the training of DIL, it is better to collect various distorted/clean image pairs with different distortions but the same content. However, in the real world, it is non-trivial to collect the datasets to satisfy this. Therefore, we can construct the ideal datasets by answering the counterfactual question \textit{``if $D$ is $d_i$, what the $I_d$ would be with $I_c$ invariant?"}. We call the construction counterfactual distortion augmentation.

Analogously,
the computing of counterfactuals in distortion augmentation follows a three-step procedure. 1) Abduction: Use the distorted image $I_d$ to determine the value of $I_c$, \textit{i.e.,} $P(I_c|I_d)$. 2) Action: Modify the degradation model, $g$, so that $D$ is \tcb{adjusted} to the counterfactual value $d_i$, that might rarely existed in real-world (\textit{e.g.,} the synthesised distortions). 3) Prediction: Compute the consequence $I_{d_i}$ of the counterfactual based on estimated $I_c$ and modified degradation model $g_{d_i}$.

It is fortunate that there are amounts of high-quality images captured by professional devices, that are only degraded by some extremely mild distortions. We can regard these images as clean images $I_c$. Therefore, the first step in counterfactuals is unnecessary and can be ignored. We can implement the counterfactual distortion augmentation by adding different synthetic distortion types or degrees to the same image contents $I_c$.

\section{The derivation of the parallel sampling.}
\label{sec:derivation}
In this section, we will give the derivation of our parallel sampling in Eq. 5 of our paper.
From Eq. 3 and Eq. 4 in our paper as:
\begin{equation}
\begin{aligned}
    \theta^* = & \arg \min_{\theta} \mathbb{E}_{(I_d, I_c)\sim \mathcal{D}}[ \frac{1}{n} \sum_{d_i\in D} \mathcal{L}(f_{\phi_{d_i}}(I_d), I_c)], \\ & where \quad
    \phi_{d_i} = \theta - \alpha \nabla_\theta \mathcal{L}(f_\theta(I_{d_i}), I_c)
    \label{equ:final-update}
\end{aligned}
\end{equation}
we can conduct the Taylor expansion for the above equation at position $\theta$ as:
\begin{equation}
    \begin{aligned}
        \theta^* &=  \arg \min_{\theta} \mathbb{E}_{(I_d, I_c)\sim \mathcal{D}}\{ \frac{1}{n}\sum_{d_i\in D}[ \mathcal{L}(f_{\theta}(I_d), I_c)  \\& \quad \quad - \alpha \nabla_\theta \mathcal{L}(f_\theta(I_{d_i}), I_c) \nabla_\theta \mathcal{L}(f_\theta(I_{d}), I_c) \\
        & \quad \quad +o(\nabla_\theta \mathcal{L}(f_\theta(I_{d}), I_c))] \} \\
        & = \arg \min_{\theta} \mathbb{E}_{(I_d, I_c)\sim \mathcal{D}} \{\mathcal{L}(f_{\theta}(I_d), I_c) \\
        & \quad \quad -\frac{1}{n}\sum_{d_i\in D}\alpha[\nabla_\theta \mathcal{L}(f_\theta(I_{d_i}), I_c)]\nabla_\theta \mathcal{L}(f_\theta(I_{d}), I_c)\\
        & \quad \quad +o(\nabla_\theta \mathcal{L}(f_\theta(I_{d}), I_c))] \}\} \\
        & = \arg \min_{\theta} \mathbb{E}_{(I_d, I_c)\sim \mathcal{D}} \{\mathcal{L}(f_{\theta}(I_d), I_c) \\
        & \quad \quad -\alpha\nabla_\theta [\sum_{d_i\in D}\frac{1}{n}\mathcal{L}(f_\theta(I_{d_i}), I_c)]\nabla_\theta \mathcal{L}(f_\theta(I_{d}), I_c)\\
        & \quad \quad +o(\nabla_\theta \mathcal{L}(f_\theta(I_{d}), I_c))] \}\} \\
    \end{aligned}
    \label{eq:4}
\end{equation}

Then we conduct the Taylor inverse expansion for the Eq.~\ref{eq:4}. The Eq.~\ref{eq:4} can be derived as:
\begin{equation}
    \begin{aligned}
        \theta^* &= 
        \arg \min_{\theta} \mathbb{E}_{(I_d, I_c)\sim \mathcal{D}}[ \\ 
        &\mathcal{L}(f_{(\theta-\alpha\nabla_\theta \sum_{d_i\in D}\frac{1}{n}\mathcal{L}(f_\theta(I_{d_i}), I_c))}(I_d), I_c)]   \\
    \end{aligned}
    \label{eq:5}
\end{equation}
Let $\phi_{\Bar{d}}= \theta - \alpha \nabla_\theta \sum_{d_i\in D}\frac{1}{n}\mathcal{L}(f_\theta(I_{d_i}), I_c)$, we can obtain the final equation as Eq. 5 of our paper as:
\begin{equation}
 \begin{aligned}
      &\theta^* = \arg\min_{\theta}  \mathbb{E}_{(I_d, I_c)\sim \mathcal{D}} [  \mathcal{L}(f_{\phi_{\Bar{d}}}(I_d), I_c)], \\ 
       & where ~\phi_{\Bar{d}}= \theta - \alpha \nabla_\theta \sum_{d_i\in D}\frac{1}{n}\mathcal{L}(f_\theta(I_{d_i}), I_c),
\end{aligned}  
\label{eq:6}
\end{equation}
\section{The detailed algorithms on four variants of DIL.} 
\label{sec:algs}
We further demonstrate the algorithm details of four variants of our proposed DIL in the Alg.~\ref{alg:ps} (DIL$_{ps}$), Alg.~\ref{alg:ss} (DIL$_{ss}$), Alg.~\ref{alg:pf} (DIL$_{pf}$), and Alg.~\ref{alg:sf} (DIL$_{sf}$). As derived in Eq.~\ref{eq:6}, we can utilize the parallel data sampling for all distortions $D$ to substitute the serial sampling based optimization. The implementation differences between the two sampling strategies can be observed by comparing the Line $5-6$ in the Alg.~\ref{alg:ps} and Line $5-9$ in the Alg.~\ref{alg:ss}. We can find that parallel sampling can reduce the number of parameter updating by $1/n$. By comparing the Alg.~\ref{alg:ps} and Alg.~\ref{alg:pf}, we can find that only first-order gradient existed in the DIL$_{pf}$, which is an approximation of the second-order optimization in Alg.~\ref{alg:ps}. The related proof can be found in the~\cite{nichol2018firstReptile}.

\begin{algorithm}[t]
\caption{DIL$_{ps}$ (The variant of DIL with parallel sampling and second-order optimization)}
\label{alg:ps}
\begin{algorithmic}[1]
\State \textbf{Input:} Training dataset $\mathcal{D}=\{I_{d_i}, I_c| 1 \le i \le n\}$, where $n$ is number of distortion types and degrees (\textit{i.e,} confounders), and $D=\{d_{i}|1 \le i \le n\}$ is the confounder set. 
\State \textbf{Init:}
Restoration network $f$ with the parameters $\theta$, 
 learning rate $\alpha$ for virtually updating, $\beta$ for the training.  
\While{not converge}
    \State Sample training pairs $(I_d, I_c)$ from $D$.
    
    \State Sample training pairs $\{I_{d_i}, I_c\}_{i=1}^{n}$  from $\mathcal{D}$.
    \State Virtual updating for the parameters $\theta$ as : 
    \Statex \quad \quad \quad $\phi_{\Bar{d}} \xleftarrow[]{} \theta - \alpha \nabla_\theta \sum_{d_i\in D}\frac{1}{n}\mathcal{L}(f_\theta(I_{d_i}), I_c)$.
    \State Updating the parameters $\theta$ with second-order gradient:
    $\theta \xleftarrow[]{} \theta - \beta \mathcal{L}(f_{\phi_{\Bar{d}}}(I_{d}), I_c)$
\EndWhile
\end{algorithmic}
\end{algorithm}

\begin{algorithm}[t]
\caption{DIL$_{ss}$ (The variant of DIL with serial sampling and second-order optimization)}
\label{alg:ss}
\begin{algorithmic}[1]
\State \textbf{Input:} Training dataset $\mathcal{D}=\{I_{d_i}, I_c| 1 \le i \le n\}$, where $n$ is number of distortion types and degrees (\textit{i.e,} confounders), and $D=\{d_{i}|1 \le i \le n\}$ is the confounder set. 
\State \textbf{Init:}
Restoration network $f$ with the parameters $\theta$, 
 learning rate $\alpha$ for virtually updating, $\beta$ for the training.  
\While{not converge}
    \State Sample training pairs $(I_d, I_c)$ from $D$.
        \For {$1 \le i \le n$}
        \State Sample training pairs $(I_{d_{i}}, I_c)$ from $\mathcal{D}$.
        \State Virtual updating for the parameters $\theta$: 
        \Statex \quad \quad \quad \quad \quad \quad $\phi_{d_{i}} \xleftarrow[]{} \theta - \alpha \nabla_{\theta}\mathcal{L}(f_{\theta}(I_{d_{i}}), I_c)$
        \State Compute the loss for the second-order gradient:  $\mathcal{L}(f_{\phi_{d_{i}}}(I_{d}), I_c)$
        \EndFor
        \State Updating the parameters $\theta$ with second-order gradient: 
         $\theta \xleftarrow[]{} \theta - \beta \frac{1}{n} \sum_{d_i\in D} \nabla_{\theta}\mathcal{L}(f_{\phi_{d_{i}}}(I_{d}), I_c)$
\EndWhile
\end{algorithmic}
\end{algorithm}

\begin{algorithm}[t]
\caption{DIL$_{pf}$ (The variant of DIL with parallel sampling and first-order optimization)}
\label{alg:pf}
\begin{algorithmic}[1]
\State \textbf{Input:} Training dataset $\mathcal{D}=\{I_{d_i}, I_c| 1 \le i \le n\}$, where $n$ is number of distortion types and degrees (\textit{i.e,} confounders), and $D=\{d_{i}|1 \le i \le n\}$ is the confounder set. 
\State \textbf{Init:}
Restoration network $f$ with the parameters $\theta$, 
 learning rate $\alpha$ for virtually updating, $\beta$ for the training.  
\While{not converge}
    \State $\Tilde{\theta}\xleftarrow[]{}\theta$
    \For {step=1 to 2}
    \State Sample training pairs $\{I_{d_i}, I_c\}_{i=1}^{n}$  from $\mathcal{D}$. 
    \State Virtual updating: 
    \Statex \quad \quad \quad \quad \quad$\Tilde{\theta} \xleftarrow[]{} \Tilde{\theta} - \alpha\nabla_{\Tilde{\theta}}\sum\limits_{d_i\in D}\dfrac{1}{n}\mathcal{L}(f_{\Tilde{\theta}}(I_{d_i}), I_c$.
    \EndFor
    \State Updating the parameters $\theta$:  $\theta \xleftarrow[]{} \theta - \beta(\Tilde{\theta}-\theta)$
\EndWhile
\end{algorithmic}
\end{algorithm}

\begin{algorithm}[t]
\caption{DIL$_{sf}$ (The variant of DIL with serial sampling and first-order optimization)}
\label{alg:sf}
\begin{algorithmic}[1]
\State \textbf{Input:} Training dataset $\mathcal{D}=\{I_{d_i}, I_c| 1 \le i \le n\}$, where $n$ is number of distortion types and degrees (\textit{i.e,} confounders), and $D=\{d_{i}|1 \le i \le n\}$ is the confounder set. 
\State \textbf{Init:}
Restoration network $f$ with the parameters $\theta$, 
 learning rate $\alpha$ for virtually updating, $\beta$ for the training.  
\While{not converge}
    \State $\Tilde{\theta}\xleftarrow[]{}\theta$
    \For {$1 \le i \le n$}
    \State Sample training pairs $(I_{d_i}, I_c)$ from $\mathcal{D}$.
    \State Virtual Updating: $\Tilde{\theta} \xleftarrow[]{} \Tilde{\theta} - \alpha\nabla_{\Tilde{\theta}}\mathcal{L}(f_{\Tilde{\theta}}(I_{d_i}), I_c)$
    \EndFor
    \State  Updating the parameters $\theta$:  $\theta \xleftarrow[]{} \theta - \beta(\Tilde{\theta}-\theta)$
\EndWhile
\end{algorithmic}
\end{algorithm}

\section{Implementation Details.}
\label{sec:implementation}
\subsection{Overall Settings.}
For all image restoration tasks (except for the image deraining task) in this paper, we use 800 images from DIV2K~\cite{DIV2K} and 2650 images from Flickr2K~\cite{Flickr2K} as the clean images to construct the datasets for training.
 Following the common setting~\cite{zamir2021multi,liang2021swinir}, In the training process, we randomly crop the distorted/clean image pairs with the size of $64\times64$ from the training images, and feed them to the restoration network to optimize the parameters. In the process of the counterfactual distortion augmentation, the distorted patches $I_d$ are generated online according to distortion set $D$ of different image restoration tasks. For ERM, we use Adam optimizer with $\beta_1=0.9$ and $\beta_2=0.999$. For DIL$_{sf}$ and DIL$_{pf}$ training paradigms, the same Adam optimizer with ERM is used for the training optimization for the above two variants. For the virtual updating process, we adopt the Adam optimizer with $\beta_1=0$ and $\beta_2=0.999$ following~\cite{nichol2018firstReptile}. 
 For DIL$_{ss}$ and DIL$_{ps}$, we utilize the same two Adam optimizers as that used in ERM for the virtual updating step and training optimization step. 
  We set the batch size to 8 on each GPU. The total training iterations and initial learning rate are set to 400K and 1e-4, respectively. The learning rate will reduce by half at [200K, 300K]. 
  All the tasks are optimized with the L1 loss if not mentioned. In the image deraining task, we utilize Charbonnier~\cite{charbonnier1994two} Loss as:
\begin{equation}
    \mathcal{L}_{char} = \sqrt{\|I_o - I_{c}\|^2 + \epsilon^2}
\end{equation}
where $I_o$ and $I_{c}$ denotes the reconstructed images and clean images, respectively. Following previous works~\cite{jiang2020multi,zamir2021multi}, we set $\epsilon$ to 1e-3.
\\
\subsection{The distortion/confounders set $D$ for different tasks.}
In this section, we describe the specific construction of the distortion/confounder set $D$ in the counterfactual distortion augmentation strategy.

\subsubsection{Cross distortion degrees}
\noindent\textbf{Image Denoising.}
For image denoising, the distortion/confounder set is composed of Additive White Gaussian Noise (AWGN) with the noise intensity of [5, 10, 15, 20], which is added to the clean images from DF2K~\cite{DIV2K,Flickr2K} to construct the training data. For testing, we utilize several unseen noise intensities, including [30, 40, 50] to estimate the generalization capability of different schemes. \\ 
\noindent\textbf{Image Deblurring.}
For image deblurring, we obtain $I_d$ by applying the distortion/confounding set $D$ to $I_c$, which contains the 2D gaussian filter with different blurring sigma of [1.0, 2.0, 3.0, 4.0]. For testing, we validate the generalization capability of different schemes on the sigma [4.2, 4.4, 4.6, 4.8, 5.0]. \\
\noindent\textbf{Hybrid distortion restoration.}
Following the~\cite{li2020learningFDRNet,yu2018craftingRLRestore}, the hybrid distortions are degraded with the superposition of blur, noise, and Jpeg compression artifacts in a sequence manner.  The distortion/confounder set $D$ for training is composed of multiple levels of severe hybrid distortions. The test datasets are composed of unseen distortion levels, including mild and moderate hybrid distortions. 

\subsubsection{Cross distortion types}

\noindent\textbf{Real Image Super-resolution.}
For real image super-resolution, we utilize the degradation model introduced by~\cite{RealESRGAN} for training. To simplify the training process, we adopt the one-order distortion synthesis mode in ~\cite{RealESRGAN} to construct the distortion/confounder set $D$, where different $d_i\in D$ are divided with different noise types and blur types in the degradation model of~\cite{RealESRGAN}. To validate the generalization capability of different schemes for the ``cross distortion types", we exploit the RealSR~\cite{RealSR} and DRealSR~\cite{DrealSR} for the real image super-resolution as our test data. 

\noindent\textbf{Real Image Denoising.}
For real image denoising, we obtain $I_d$ based on the ISP process introduced by~\cite{CBDNet}. We divide this degradation model into four different distortion types based on the different color filter arrays (CFA) to construct $D$. We follow previous works~\cite{DND,SIDD} and utilize real denoising datasets DND~\cite{DND} and SIDD~\cite{SIDD} as the benchmarks to validate the generalization capability of different schemes.\\
\noindent\textbf{Image Deraining.}
For image deraining, we utilize three datasets with three different raining types, including  Rain14000~\cite{Rain12600}, DID-MDN~\cite{DID-MDN}, and Heavy Rain Dataset~\cite{heavyrain}, to construct the training data $\mathcal{D}$, and test the generalization capability of the restoration network on other three unseen raining types, including 100 rainy images from Rain100L~\cite{Rain100}, 100 rainy images from Rain800~\cite{Rain800}, and 12 rainy images from Rain12~\cite{Rain12}. It is noteworthy that the synthesis strategies of the above raining types are rarely released. Therefore, in this task, we relax the content consistency between different raining types for training. And our DIL is still effective for improving the generalization capability of the restoration network.



\begin{figure*}[ht]
    \centering
    \includegraphics[width=0.95\linewidth]{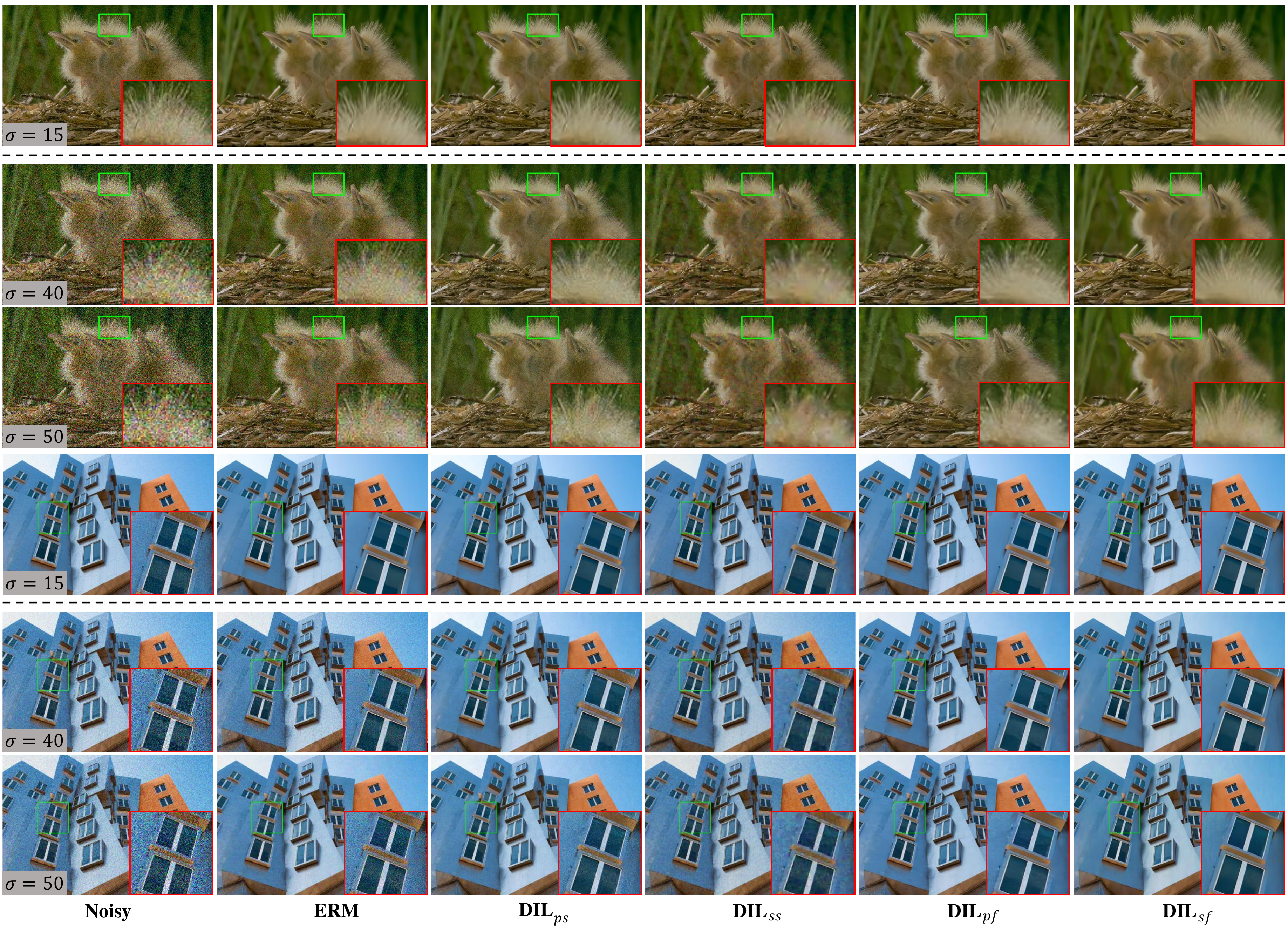}
    \caption{Visual comparison of the commonly-used ERM and our proposed four variants of DIL on image denoising task. The noise level above the dash line is seen, while noise levels below are unseen.}
    \label{fig:noise_supp}
    
\end{figure*}

\begin{figure*}[ht]
    \centering
    \includegraphics[width=0.95\linewidth]{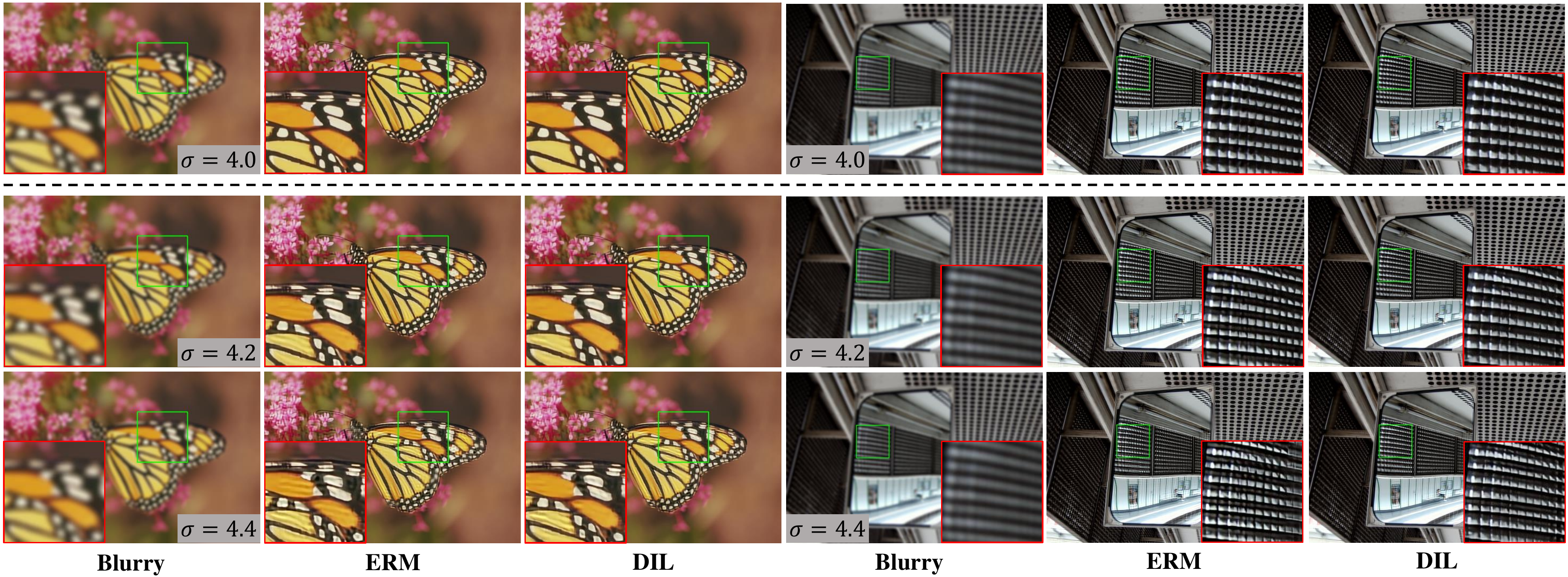}
    \caption{Visual comparison of the commonly-used ERM and DIL on image deblurring task. The blur level above the dash line is seen, while blur levels below are unseen.}
    \label{fig:blur_supp}
    
\end{figure*}

\begin{figure*}[ht]
    \centering
    \includegraphics[width=0.95\linewidth]{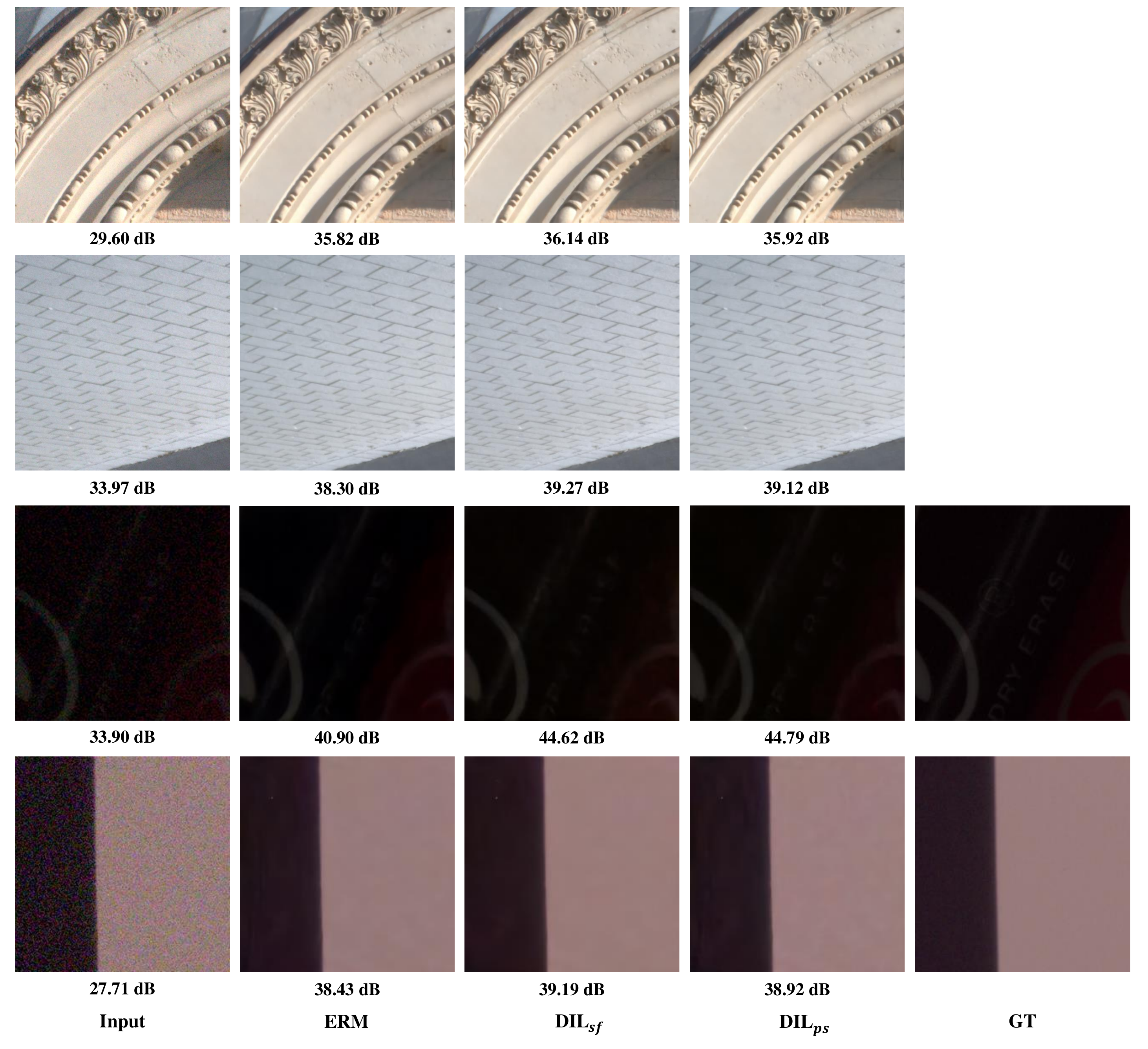}
    \caption{Visual comparison of the commonly-used ERM and DIL on real image denoising task. The top samples are from DND~\cite{DND} while the bottom samples are from SIDD~\cite{SIDD}. Brightening the third line for a better view.}
    \label{fig:realnoise_supp}
    
\end{figure*}

\begin{figure*}[ht]
    \centering
    \includegraphics[width=0.95\linewidth]{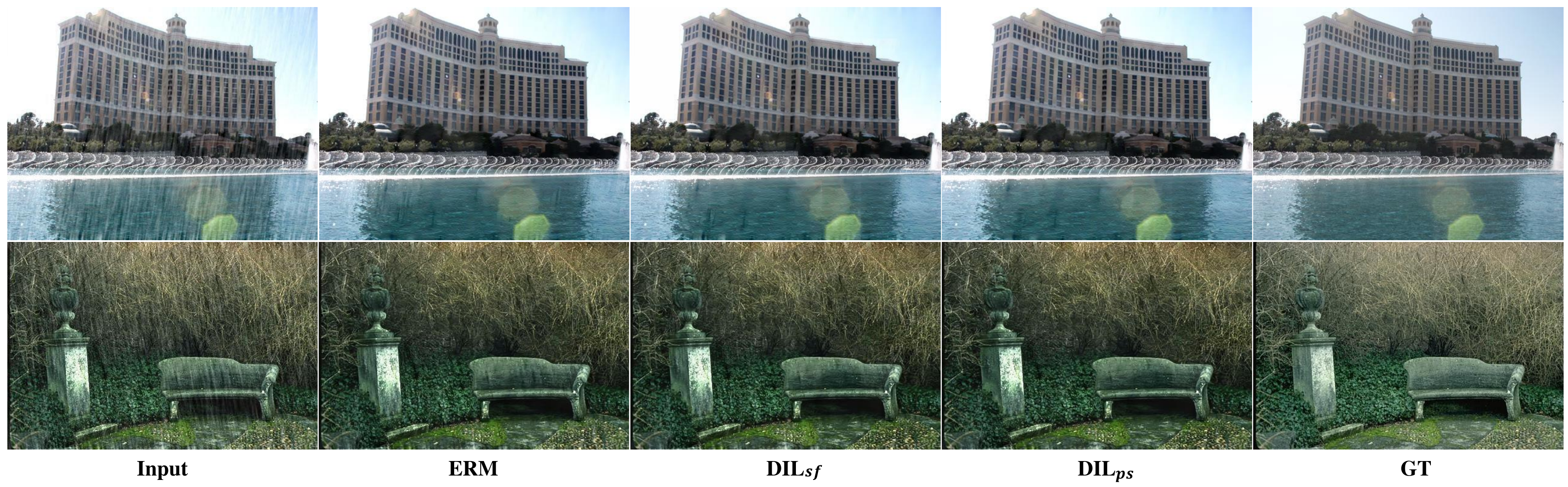}
    \caption{Visual comparison of the commonly-used ERM, our DIL$_{sf}$ and DIL$_{ps}$ on image deraining task.}
    \label{fig:derain_supp}
    
\end{figure*}

\begin{figure*}[ht]
    \centering
    \includegraphics[width=0.95\linewidth]{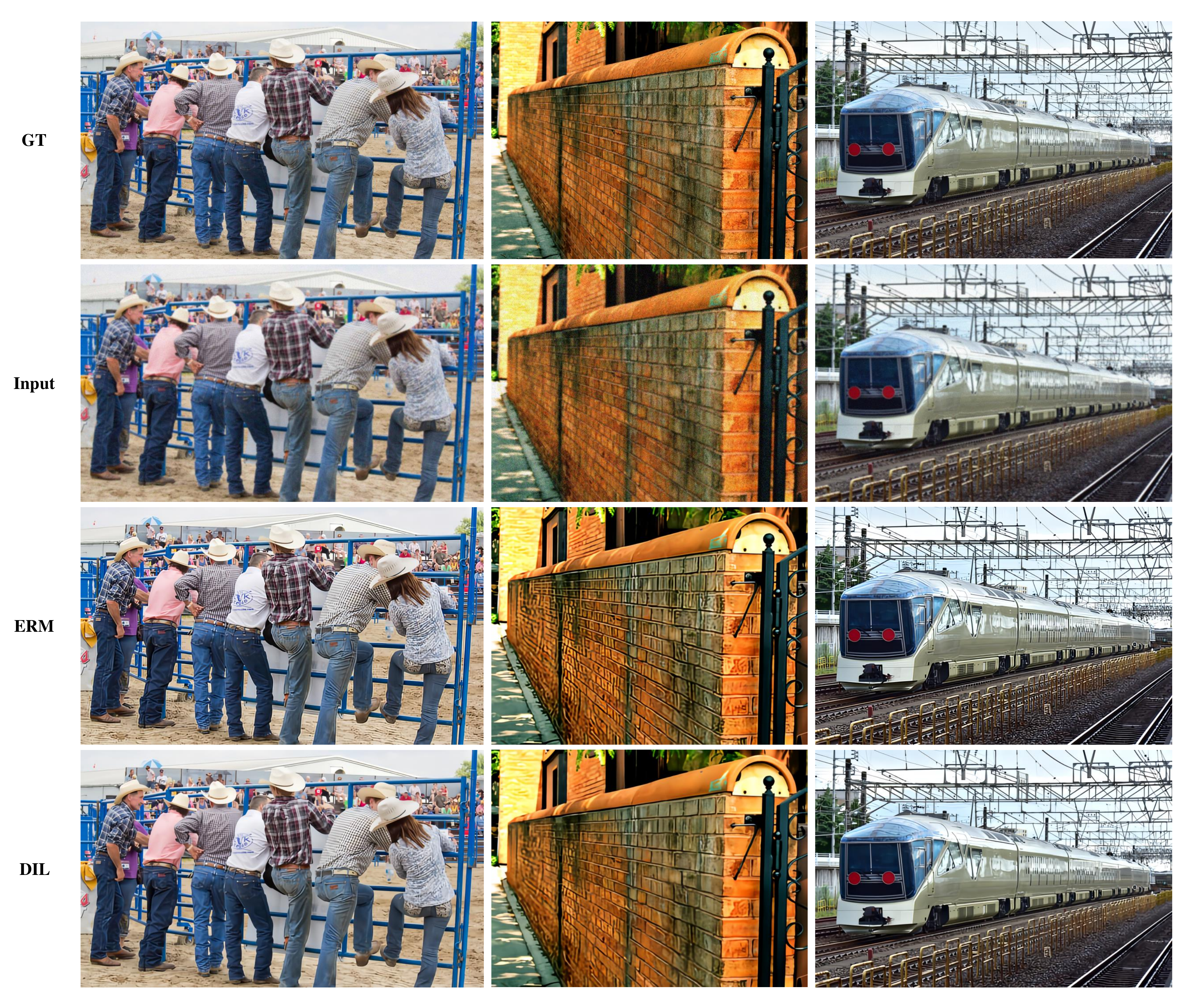}
    \caption{Visual comparison of the commonly-used ERM and DIL on hybrid distortion removal task. Here, we show restoration results on the mild distortion level, which is the unseen distortion level for the restoration network.}
    \label{fig:hybrid_supp}
    
\end{figure*}

\section{More Subjective Visualization}
\label{sec:subjective}
We provide more visual comparisons for different image restoration tasks in this section. As shown in Fig.~\ref{fig:noise_supp}, the commonly-use ERM and our proposed DIL all achieve similar reconstructed results on the seen noise level ($\sigma=15$). But ERM fails to restore high-quality images on unseen noise levels well, (\textit{e.g.,} $\sigma=40$ and $\sigma=50$), which indicates that ERM lacks of enough generalization ability for the unseen distortion degrees. In contrast,  our DIL$_{sf}$ can recover noise-free and structure-preserved images despite the distortion degrees do not exist in the training data. This further proves the correctness and effectiveness of our proposed DIL. 

We show more visual comparisons of image deblurring in Fig.~\ref{fig:blur_supp}. When dealing with unseen blur degrees, our proposed DIL can restore the clear structures, while ERM produces overshooting artifacts on the edges. More visualizations for real image denoising can be found in Fig.~\ref{fig:realnoise_supp}. And more visualizations for image deraining can be found in Fig.~\ref{fig:derain_supp}. We also visualize the subjective comparisons on hybrid-distorted image restoration in Fig.~\ref{fig:hybrid_supp}.

\end{document}